\documentclass[final,5p,times,twocolumn,numbers]{elsarticle}

\makeatletter
\def\ps@pprintTitle{%
  \let\@oddhead\@empty
  \let\@evenhead\@empty
  \def\@oddfoot{}%
  \let\@evenfoot\@oddfoot}
\makeatother

\usepackage{amssymb}
\usepackage{graphicx}
\usepackage{epstopdf}
\usepackage{epsfig}
\usepackage{lipsum}
\usepackage{subcaption}
\usepackage{amsmath}
\usepackage{hyperref}
\usepackage{multirow}
\usepackage[table,xcdraw]{xcolor}
\usepackage{colortbl}
\usepackage[normalem]{ulem}
\usepackage{array}
\usepackage{float}
\usepackage{url}
\usepackage{caption}
\usepackage{subcaption}
\usepackage{natbib}
\usepackage{pifont}
\usepackage{array} 
\usepackage{booktabs}
\useunder{\uline}{\ul}{}

\hypersetup{
    colorlinks=true,
    linkcolor=blue,
    filecolor=magenta,
    urlcolor=cyan,
}

\bibpunct{[}{]}{;}{n}{,}{,}

\begin{document}

\begin{frontmatter}

\title{\textbf{\LARGE Novel Deep Learning Architectures for Classification and Segmentation of Brain Tumors from MRI Images}}

\author[1]{\Large Sayan Das\corref{equal}}
\author[2]{\Large Arghadip Biswas\corref{equal}}

\affiliation[1]{organization={IIIT Delhi}, city={Delhi}, country={India}}

\affiliation[2]{organization={Jadavpur University}, 
            city={Kolkata},
            country={India}}

\cortext[equal]{These authors contributed equally to this work.}

\cortext[note]{\textit{Note: This is a preprint made available via arXiv for open access and academic dissemination. It has not been submitted for peer-reviewed journal publication.}}

\begin{abstract}
\label{sec:abstract}

Brain tumors pose a significant threat to human life, therefore it is very much necessary to detect them accurately in the early stages for better diagnosis and treatment. Brain tumors can be detected by the radiologist manually from the MRI scan images of the patients. However, the incidence of brain tumors has risen amongst children and adolescents in recent years, resulting in a substantial volume of data, as a result, it is time-consuming and difficult to detect manually. With the emergence of Artificial intelligence in the modern world and its vast application in the medical field, we can make an approach to the CAD (Computer Aided Diagnosis) system for the early detection of Brain tumors automatically. All the existing models for this task are not completely generalized and perform poorly on the validation data. So, we have proposed two novel Deep Learning Architectures - (a) SAETCN (Self-Attention Enhancement Tumor Classification Network) for the classification of different kinds of brain tumors. We have achieved an accuracy of 99.38\% on the validation dataset making it one of the few Novel Deep learning-based architecture that is capable of detecting brain tumors accurately. We have trained the model on the dataset, which contains images of 3 types of tumors (glioma, meningioma, and pituitary tumors) and non-tumor cases. and (b) SAS-Net (Self-Attentive Segmentation Network) for the accurate segmentation of brain tumors. We have achieved an overall pixel accuracy of 99.23\%.
\end{abstract}

\begin{keyword}
Brain Tumor \sep MRI Images \sep Deep Learning \sep Machine Learning \sep CNN \sep Classification
\end{keyword}

\end{frontmatter}

\section{Introduction}
\label{sec:intro}

\begin{figure}
    \centering
    \includegraphics[width=0.4\textwidth]{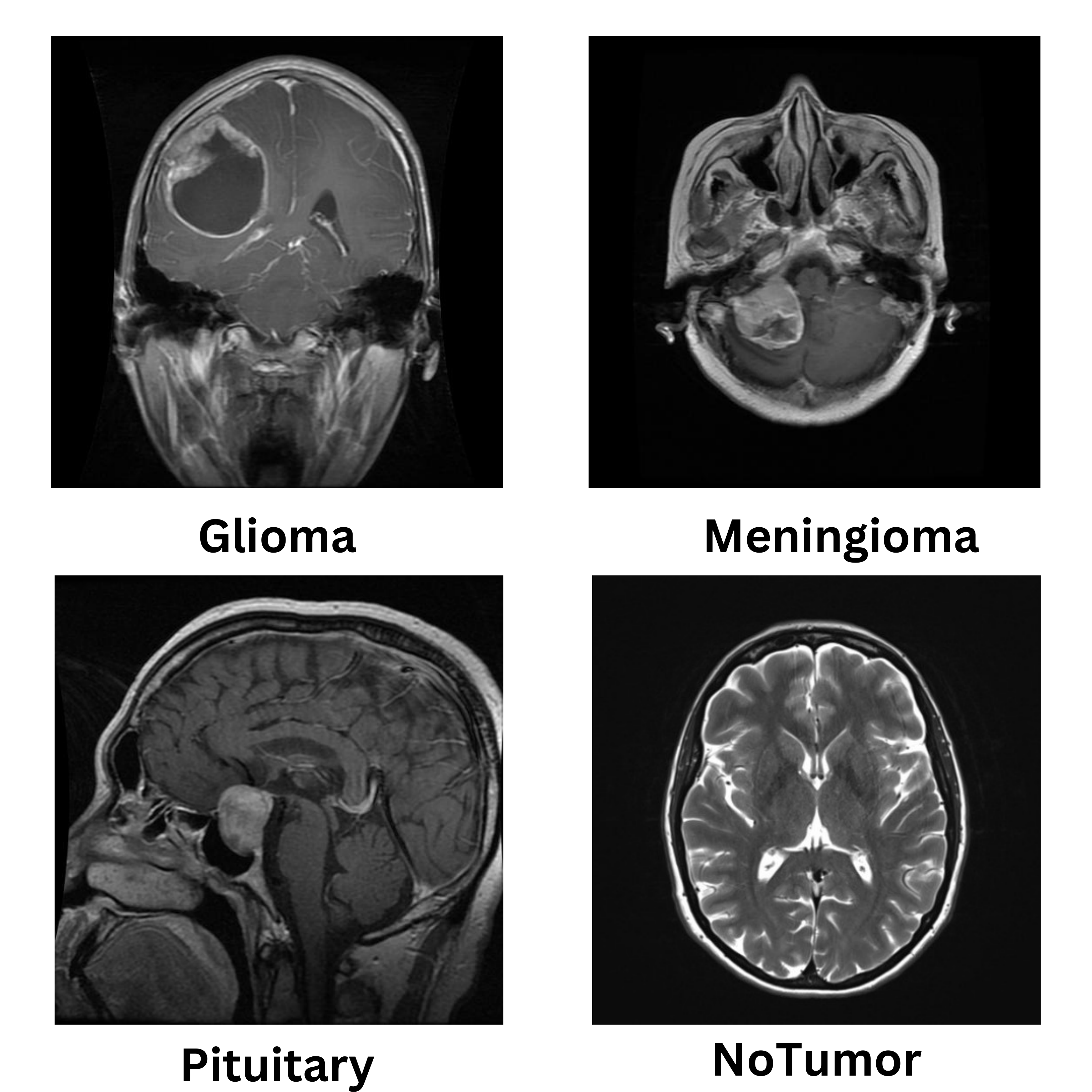}
    \caption{MRI Images of Types of Brain Tumor and No Tumor}
    \label{fig: glioma}
\end{figure}

Brain Tumors are a huge concern in the field of medicine because of their high mortality rate. Brain tumor forms when there is an uncontrollable abnormal growth of the cells within the Brain. The abnormal growth may occur in the brain itself which is called a primary tumor or it may spread to the brain from the other parts of the body which are called secondary or metastatic tumors \cite{i4}. The proper reason and causes of brain tumors are not yet understood but according to researchers, they occur due to genetic mutations that affect cell growth and division \cite{i5}. This mutation can cause the cell to multiply causing the tumor. There are 120 types of tumors out of which some tumors are benign (non-cancerous) which grows slowly and some are malignant (cancerous) which grow rapidly \cite{i6}. These tumors pose a significant global health challenge due to their high mortality rates, and complex procedure of treatment. Some common types of Brain tumors that we are going to detect and classify from the MRI Images by our proposed novel deep learning architectures - SAETCN and SAS-Net are - Gliomas, meningiomas, and Pituitary. All the brain tumor types and an MRI image with no tumor are shown in Figure \ref{fig: glioma}.

Gliomas are the most common type of Brian tumors. Around 30\% of all brain and CNS tumors and 80\% of all malignant tumors are of this kind. This type of tumor can occur at any age, but they are most common in the age range between 45 and 65 years of age. The incidence of gliomas has been stable in the last few years but due to the advancement of treatment, the mortality rate slightly increased \cite{i1}.

Meningiomas are also one of the common primary types of brain tumors. Around 37\% of all brain tumors are of this kind. They are mostly common amongst older individuals with ages greater than 65. The incidence of occurrence of this tumor is increasing and they are benign but their location can make them difficult to treat\cite{i1}.

Around 15\% of all brain tumors are of type Pituitary. This can occur at any age group but the most common age range is between 30 to 50 years. \cite{i2} Most of them are benign in nature and are slow-growing. The detection of pituitary tumors has increased due to the MRI images.

The overall rate of occurrence of brain tumors is 6.2 per 1,00,000 people per year and the death rate is 4.4 per 1,00,000 people per year \cite{i1}. Brain tumors are the leading cause of cancer-related deaths in males aged birth to 39 years and females aged birth to 19 years \cite{i3}. There are almost 2,08,620 cases of brain tumors in adolescents and adults in the age group of 19 to 39, who are living with a primary brain or spinal cord tumor in the United States. The pituitary tumors are most common among those incidents in that age group \cite{i2}. The 5-year relative survival rate for brain and other nervous system cancers is about 33.4\% \cite{i1}.

\begin{figure*}
    \centering
    \includegraphics[width=0.85\textwidth]{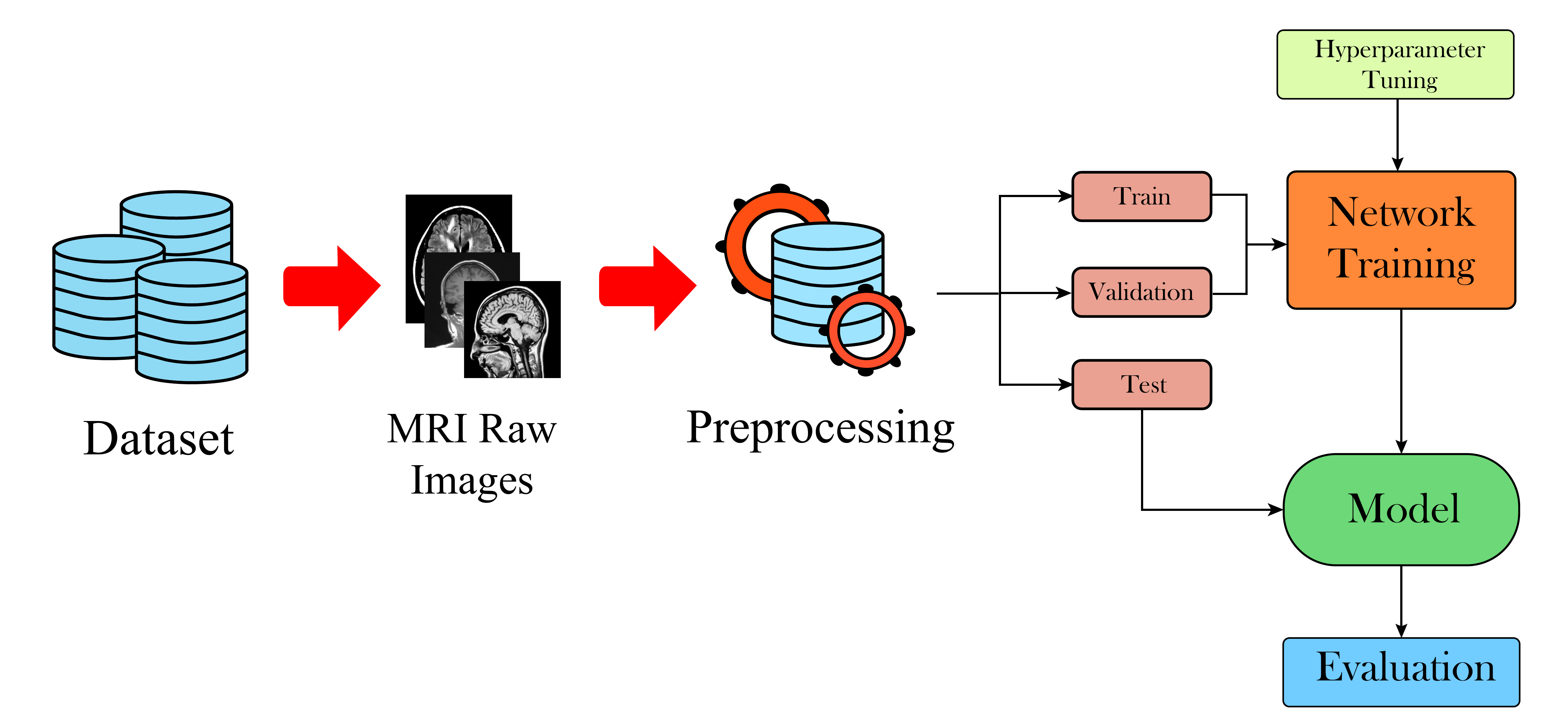}
    \caption{Processflow Diagram of our Experiment}
    \label{fig: processflow}
\end{figure*}

The brain is the most important organ of our body, Proper functioning of the brain is very crucial for overall body performance. So, Both benign and malignant tumors present in the brain pose a significant challenge due to their complex nature and the critical functions of the brain. The accurate classification and segmentation of different types of brain tumors, such as gliomas, meningiomas, and pituitary tumors, in early stages, is essential for effective treatment of the patient. The advancement of Artificial Intelligence in the modern world and its vast application in the field of medicine help us to classify and segment different types of tumors and their exact position. Over the last few years, several researches have taken place to support that deep learning approaches are one of the best for the classification of brain tumors automatically and efficiently with a good accuracy rate and fewer errors \cite{ rw2, rw7, rw4, rw1}. This paper consists of a detailed explanation of our novel proposed deep learning architecture with four modules that are made with 16 layers of custom-made SAE blocks with attention and skip connection mechanism inspired from the residual and inception operations divided into four modules, giving a good accuracy that beats any state of the art deep learning ImageNet architectures. The study focuses on enhancing the accuracy and efficiency of brain tumor classification and early detection along with proper segmentation, which is crucial for effective diagnosis and treatment planning. The research contribution of this study is as follows:

\begin{itemize}
    \item Data Preprocessing: Effective data preprocessing has been done from the figshare data which was on mat files, they are converted into image files. Basic preprocessing like contrast enhancement and normalization has been done for the other two datasets. Both three and four-class classifications have been performed based on our dataset.
    \item The proposition of our Novel Architecture for Classification: SAETCN or Self-Attention Enhanced Tumor Classification Network is a completely novel architecture that has the power to beat many state-of-the-art deep learning models for the classification of the brain tumor into three classes - glioma, meningioma, and pituitary from the MRI images.
    \item The proposition of our Novel Architecture for Segmentation: SAS-Net or Self Attentive Segmentation Network is a completely novel deep learning architecture that can segment the tumour region accurately with an overall pixel accuracy of 99.23\%.
\end{itemize}

In this paper, we have discussed the following sections: Section \ref{sec:related works} discusses the related works that have been done in this field and the recent state-of-the-art methods and their outcomes. Section \ref{sec: proposed methodology} explains our proposed methodology for the classification and segmentation of brain tumors. Section \ref{sec:experiment} provides the experimental setup, details about the dataset utilized in this study, and the experimental results obtained through the methodology. Section \ref{sec:future scope} discusses the future scope of this research and our plan to integrate this model into a mobile application. Section \ref{sec:conclusions} discusses the conclusions drawn from the study.

\section{Related Works}
\label{sec:related works}

Over the last few years, many researches have been done and many methodologies have been proposed for the detection, classification, and segmentation of brain tumors from MRI images.

\subsection{Review of previous works on classification}
Muhammad Aamir \cite{rw5} used an optimized convolutional neural network on the Brain MRI dataset present in Kaggle and got an accuracy of 97\%. However the limitation of this model is that it is not completely generalized and as a result, the model gets overfitted to the training data and may work poorer in the unseen data despite having good performance metrics. 

Saeedi et al. proposed a 2D CNN and an auto-encoder for the classification of brain images \cite{rw1}. They have used a dataset comprised of 3264 T1-weighted contrast-enhanced MRI images. The training accuracy of the proposed 2D CNN and that of the proposed auto-encoder network were found to be 96.47\% and 95.63\%, respectively and the test accuracies are 93.44\% and 90.92\%. The drawback of this paper is that the test accuracy is quite low and thus it may hinder the accurate prediction and classification of the brain tumor types in real-life scenarios thus affecting the treatment and diagnosis procedure.

Takowa Rahman \cite{rw4} used PDCNN over three datasets - The binary tumor identification dataset-I, Figshare dataset-II, and Multiclass Kaggle dataset-III provided accuracy of 97.33\%, 97.60\%, and 98.12\%, respectively on the augmented dataset. But in the original dataset, they got a result of 96\%, 96.10\%, and 95.60\% respectively. It is clear that the model gave better results on the augmented data than that of the original data. The major drawback of using augmented data is that it may introduce some artifacts that do not present in the real MRI Images. Also, the model may get overfitted with the augmented data, This can lead to models that perform well on augmented data but poorly on real, unseen data. 


Badža and Barjaktarović conducted a study utilizing a convolutional neural network (CNN) to classify glioma, meningioma, and pituitary tumors \cite{rw7}. The architecture of the network included an input layer, two “A” blocks, two “B” blocks, a classification block, and an output layer, totaling 22 layers. The performance of the network was assessed using the k-fold cross-validation method, achieving a peak accuracy of 96.56\% with tenfold cross-validation. The dataset comprised 3064 T1-weighted contrast-enhanced MRI images sourced from Nanfang Hospital, General Hospital, and Tianjin Medical University in China. 

Nyoman Abiwinanda \cite{rw2} used a CNN to classify the three most common types of brain tumors: Gliomas, Meningiomas, and Pituitary. In the study, the CNN was trained in the 2017 figshare dataset which contains 3064 T-1 weighted CE-MRI from brain tumor images provided by Cheng. He used an "Adam" optimizer and 32 filters were applied in all the convolutional layers with a ReLU activation function and a max pool with size 2 x 2, and a fully connected layer of 64 neurons with a softmax activation function. The best-reported accuracy rates for training and validation were 98.51\% and 84.19\%, respectively. The low test accuracy is a major limitation of this model that doubts the real-life classification of brain tumors.

Sai Samarth R. Phaye \cite{rw6} developed capsule algorithm networks. (DCNet) and diverse capsule networks (DCNet++). A deeper convolutional network was added in DCNet to learn distinctive feature maps. The DCNet is more efficient for learning complex data. They used a dataset comprising 3064 MRI images of 233 brain tumor patients for classification. The accuracy of the DCNet algorithm test was 93.04\%, and the accuracy of the DCNet++ algorithm was 95.03\%.

\begin{figure*}[ht]
    \centering
    \includegraphics[width=\textwidth]{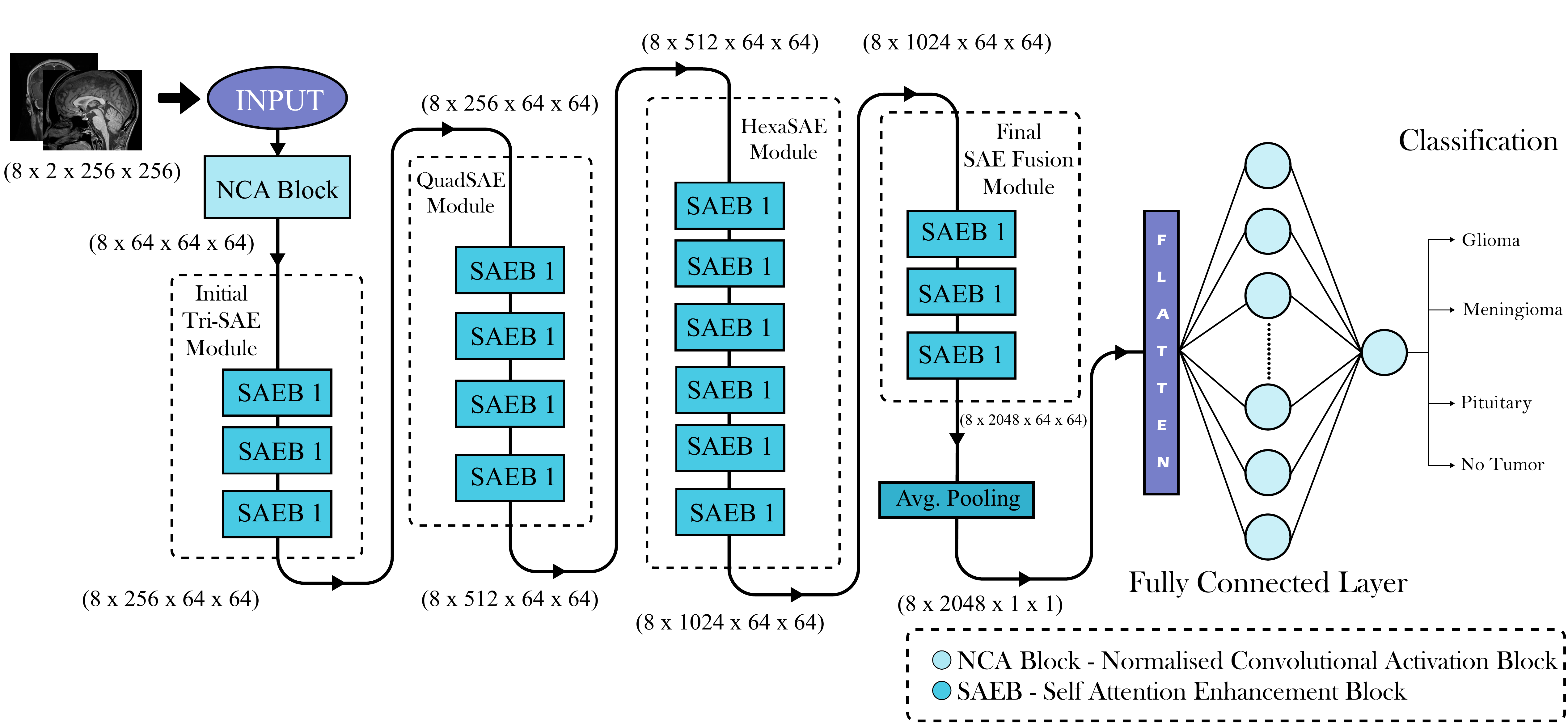}
    \caption{Self-Attention Enhanced Tumor Classification Network (SAETCN)}
    \label{fig: saetcn}
\end{figure*}

\begin{figure*}[ht]
    \centering
    \includegraphics[width=0.9\textwidth]{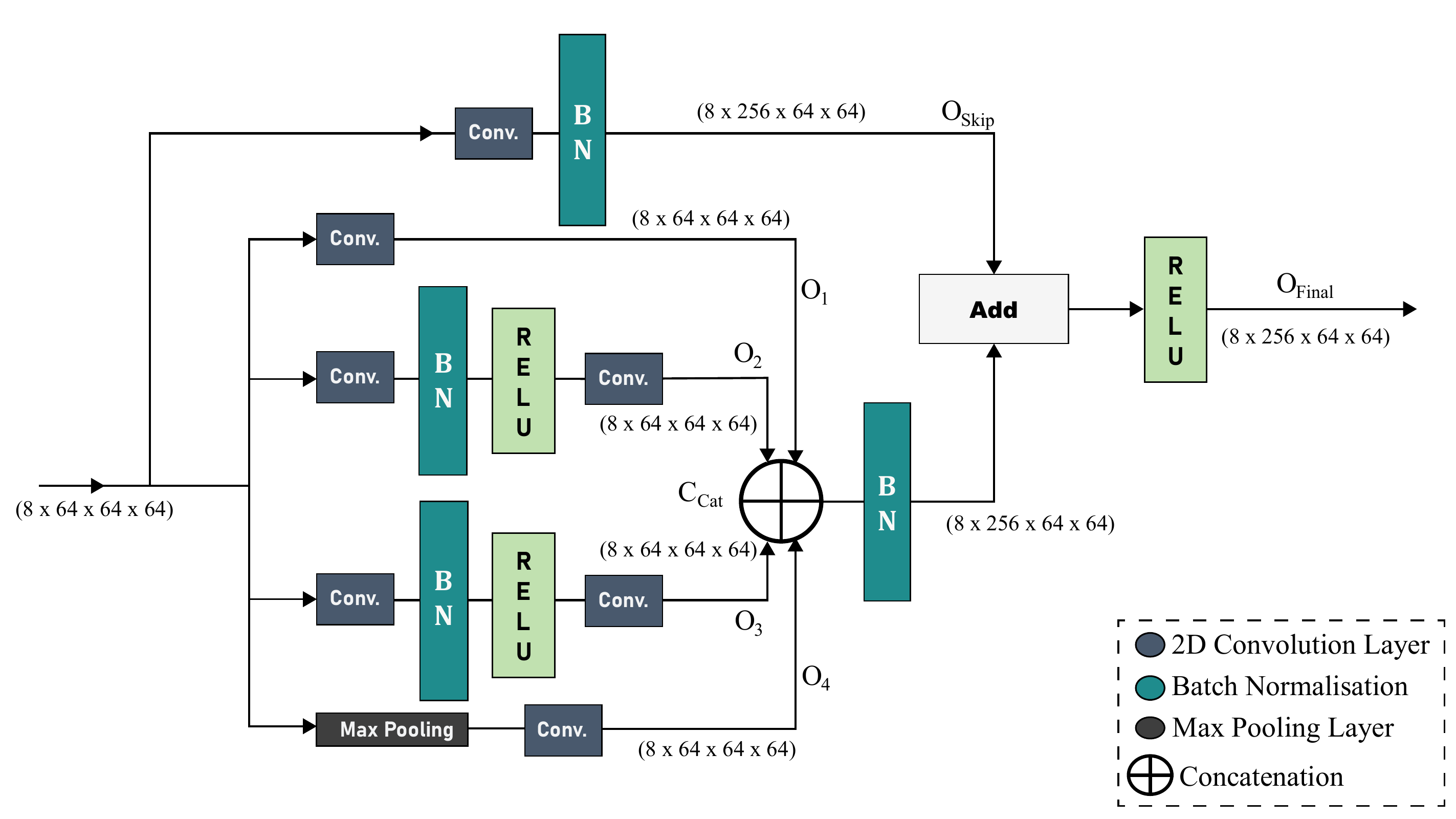}
    \caption{Self Attention Enhancement Block}
    \label{fig: saeblock}
\end{figure*}

Pashaei et al. offered various methods for identifying three types of tumors: meningiomas, gliomas, and pituitary malignancies \cite{rw8} . To extract information from photographs, the technique employed a convolutional neural network (CNN). The architecture had four convolutional layers, four pooling layers, one fully connected layer, and four batch normalization layers. The model was trained in 10 epochs, each with 16 iterations and a learning rate of 0.01. Cheng contributed to the data utilized in this study. The model's performance was evaluated using a tenfold cross-validation technique, with 70\% of the data utilized for training and 30\% for testing. The suggested algorithm outperformed MLP, Stacking, XGBoost, SVM, and RBF in terms of accuracy, with a 93.68\% success rate. 

Paul et al. \cite{rw9} used deep learning algorithms to categorize brain scans of meningiomas, gliomas, and pituitary cancers. They analyzed 3064 T1-weighted contrast-enhanced MRI images from 233 individuals. The study developed two types of neural networks: fully connected networks and CNNs. The fivefold cross-validation procedure demonstrated that generic techniques performed better than particular methods that needed picture dilation, with an accuracy of 91.43\%. 

In all those previous studies we have found some major drawbacks. In \cite{rw5} the model is not generalized enough. In \cite{rw1, rw2} the model performs poorly on the test dataset and thus becomes risky to implement it in real world. In \cite{rw4} the model performs good in augmented data but not on the original data.

\subsection{Review of previous works on segmentation}
Brain tumor segmentation has been a focal point of research in medical imaging, with deep learning methods showing significant promise. Liu et al.\cite{liu2023deep} conducted a comprehensive survey on deep learning-based brain tumor segmentation, highlighting various network architectures, segmentation under imbalanced conditions, and multi-modality processes. Their work underscores the importance of robust models capable of handling diverse imaging conditions. Aggarwal et al.\cite{aggarwal2023early} proposed an improved residual network (ResNet) for brain tumor segmentation, addressing gradient diffusion issues and achieving higher precision compared to traditional methods. Their approach demonstrates the potential of enhanced deep learning models in improving segmentation accuracy. Bouhafra and El Bahi\cite{bouhafra2024deep} reviewed deep learning approaches for brain tumor detection and classification, emphasizing the role of transfer learning, autoencoders, and attention mechanisms. Their analysis provides valuable insights into the advancements and limitations of current methodologies. Rao and Karunakara\cite{rao2021comprehensive} presented a comprehensive review on MRI-based brain tumor segmentation, discussing semi-automatic techniques and the latest trends in deep learning methods. Their work highlights the ongoing evolution of segmentation techniques and the need for continuous improvement. Verma et al.\cite{verma2024comprehensive} conducted a comparative study of brain tumor segmentation methods, categorizing them into conventional, machine learning-based, and deep learning-based approaches. Their findings indicate that deep learning models, particularly variants of the U-Net, outperform other methods in terms of accuracy and reliability. These studies collectively contribute to the understanding and advancement of brain tumor segmentation, providing a foundation for the development of novel deep-learning architectures.

\section{Proposed Methodology}
\label{sec: proposed methodology}

\subsection{Overall Framework}
In this study, we present a comprehensive framework that addresses both the classification and segmentation of brain tumors, leveraging advanced neural network architectures to enhance performance across these critical tasks. For tumor classification, we propose the Self Attention Enhancement Tumor Classification Network (SAETCN), designed to improve the accuracy of distinguishing between different types of brain tumors. SAETCN utilizes self-attention mechanisms to focus on relevant features within the medical imaging data, effectively capturing intricate patterns that differentiate tumor types. On the segmentation front, we introduce the Self-Attentive Segmentation Network (SAS-Net), a robust model aimed at precisely delineating tumor boundaries within the brain. SAS-Net employs similar self-attentive techniques to SAETCN, ensuring a consistent and high-quality extraction of spatial features necessary for accurate tumor segmentation. Together, these models form a unified framework that not only enhances the classification and segmentation tasks individually but also paves the way for integrated approaches that can significantly improve the overall accuracy and reliability of brain tumor diagnosis.

\subsection{Classification Architecture}
We have proposed a novel deep-learning architecture - Self-Attention Enhanced Tumor Classification Network(SAETCN) for the classification of Brain tumors. Our model has beaten many state-of-the-art deep learning architectures by achieving an accuracy of 99.31\% and 98.20\% in the Brain MRI dataset from Kaggel and Figshare Dataset by Jun Cheng respectively.

A multiclass classification of four types of classes is performed by our proposed architecture - SAETCN which is inspired by the Residual and Inception theory. The detailed architecture and mechanism of our proposed model are described in the following subsections.

\subsubsection{SAETCN Architecture}
Our proposed Deep Learning CNN architecture - SAETCN stands for Self-Attention Enhanced Tumor Classification Network, which can accurately classify 3 classes of tumor and non-tumor patients from the MRI images given as input into the Model. As shown in Figure \ref{fig: saetcn} SAETCN consists of five main parts: "Normalised Convolutional Activation Block", "Initial TriSAE Module", "QuadSAE Module", "HexaSAE Module" and "Final SAE-Fusion Module". Each Module Consists of several Self Attention Enhancement Blocks. The SAETCN Architecture has 16 custom-made Self Attention Enhancement Blocks arranged serially one after another. After the Complete Feature Extraction, the feature maps go through the Adaptive Averagepooling Layer and finally into the Dense Layer for non-linear classification.

\subsubsection{Normalised Convolutional Activation Block}
This is the initial Block of this novel Architecture. The main operation of the Normalised Convolutional Activation Block (NCAB) is the primary extraction of underlying spatial features from the input images, detecting the important textures and edges. The input images are converted into tensors and then directly sent into the NCAB as input to the Network. The NCAB consists of a Convolutional layer with the size $7\times$7, with ReLu activation followed by a Batchnormalisation and a Max Pooling Layer for dimensionality reduction. \\

Let the output result of the NCAB be denoted as \( F_{\text{NCAB}} \). The process at this stage can be expressed as:

\begin{equation}
F_{\text{NCAB}} = K_{\text{NCAB}}(X)
\end{equation}

Where, \( F_{\text{NCAB}} \) is the output feature map of the Neural Convolutional Attention Block (NCAB).
\( K_{\text{NCAB}} \) is the function representing the operations within the NCAB, which includes the convolutional layer, ReLU activation, batch normalization, and max pooling.
\( X \) is the input feature map to the NCAB. The internal operations of the \( K_{\text{NCAB}} \) is represented further in the equation \ref{eq:NCAB}.

\begin{equation}
    \begin{split}
        K_{\text{NCAB}}(X) = \text{MaxPool}(\text{BatchNorm}(\text{ReLU} \\
        (\text{Conv2D}(X_{\text{input}}, \text{kernel size}=7 \times 7))))
    \end{split}
    \label{eq:NCAB}
\end{equation}

\begin{figure}[H]
    \centering
    \includegraphics[width=0.5\textwidth]{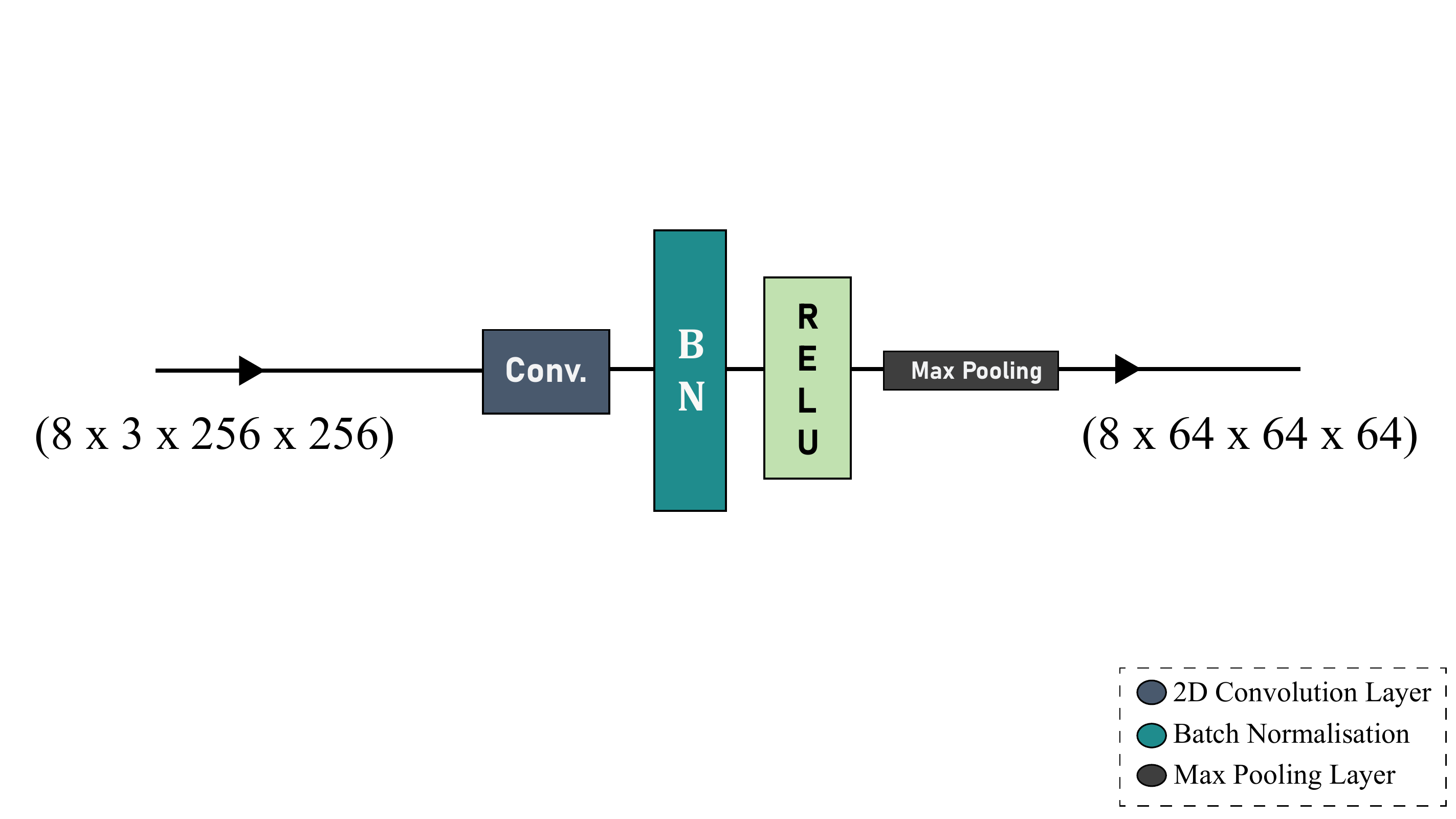}
    \caption{Normalised Convolutional Activation Block}
    \label{fig: ncab}
\end{figure}

\subsubsection{Self Attention Enhancement Block}
The Self Attention Enhancement Block (SAEB) is the basic Fundamental Block that is used in our proposed architecture. This architecture is inspired by the operation of residual and inception blocks. The basic operations of both of these blocks are described below:

\begin{itemize}
    \item Residual Blocks use the concept of skip connections that allow the gradient to flow directly and help in training deep networks.
    The residual block can be mathematically represented as: 
    \[
        y = F(x, \{W_i\}) + x
    \] 
    Where, \( x \) is the Input to the block, \( F \) is the Residual function (e.g., convolution, batch normalization, ReLU), and \( \{W_i\} \) are the Weights of the layers within the block.

    \item Spatial Features are captured at multiple scales by the inception operation. It does this by applying convolutions of different kernel sizes in parallel and concatenating their outputs.
    The Inception operation can be mathematically represented as: 
    Mathematically, the output \( O \) of an Inception module can be expressed as:

    \[
    O = O_1 \oplus O_3 \oplus O_5 \oplus O_{\text{pool}}
    \]

    Where, 
    \( O_1  \) is the Output from \(1 \times 1\) convolution branch, \( O_3  \) is the Output from \(3 \times 3\) convolution branch, \( O_5  \) is the Output from \(5 \times 5\) convolution branch, and
    \( O_{\text{pool}} \) is the Output from the pooling branch.

    Here, \(\oplus\) denotes the concatenation operator.
\end{itemize}

Using the concept of both of these operations in our Self Attention Enhancement Block (SAEB) has benefitted a lot. The skip connection helped to pass the gradient directly while skipping some layers making the deep network train faster and the inception operation helped to capture multiple spatial features from the image. As shown in Figure \ref{fig: saeblock}, the input of the SAEB block is divided into 4 Branches and a skip connection line. The first branch goes through a single Convolutional Layer. The second Branch goes through the convolutional layer with batch normalization and ReLu activation function followed by another convolutional layer with size $3\times3$. The third branch consists of the same layers as that of the second branch but the size of the convolutional layer is $5\times5$. The fourth branch goes through a max pooling with size $3\times3$ followed by a convolutional layer. The skip connection line from the input goes through a convolution layer with batch normalization. The process can be expressed as follows:\\

The output of each branch is denoted as \( O_i \) (where \( i = 1, 2, 3, 4 \)), and the output of the skip connection line is denoted as \( O_{\text{skip}} \).

\begin{align}
    O_1 &= f_{\text{conv}}(X) \\
    O_2 &= f_{\text{conv3}}(f_{\text{BN}}(f_{\text{ReLU}}(f_{\text{conv}}(X)))) \\
    O_3 &= f_{\text{conv5}}(f_{\text{BN}}(f_{\text{ReLU}}(f_{\text{conv}}(X)))) \\
    O_4 &= f_{\text{conv}}(f_{\text{pool}}(X)) \\
    O_{\text{skip}} &= f_{\text{BN}}(f_{\text{conv}}(X))
\end{align}

Where,
\( X \) is the Input feature map to the SAEB,
\( O_i \) is the Output of the \(i\)-th branch,
\( O_{\text{skip}} \) is the Output of the skip connection line,
\( f_{\text{conv}} \) is Convolutional layer,
\( f_{\text{conv3}} \) is Convolutional layer with kernel size \( 3 \times 3 \),
\( f_{\text{conv5}} \) is the Convolutional layer with kernel size \( 5 \times 5 \),
\( f_{\text{BN}} \) is the Batch normalization,
\( f_{\text{ReLU}} \) is the ReLU activation function,
\( f_{\text{pool}} \) is the Max pooling layer with size \( 3 \times 3 \).

All the outputs of the first four branches are concatenated and a batch normalization operation is done on the resulting output. This process can be expressed as:

\begin{equation}
O_{\text{concat}} = \text{BN}(O_1 \oplus O_2 \oplus O_3 \oplus O_4)
\end{equation}

where \(\oplus\) denotes the concatenation operator.

Finally, the output is added with the skip connection layer and an activation function ReLu is activated to the ultimate output.

\begin{equation}
O_{\text{final}} = \text{ReLU}(O_{\text{concat}} + O_{\text{skip}})
\end{equation}

\subsubsection{Initial TriSAE Module}
The Initial TriSAE Module is the Primary Module of the Architecture that contains three Self-Attention Enhancement Blocks (SAEB). It has 64 input channels and 256 output channels. The result \( F_{\text{NCAB}} \) of the Normalised Convolutional Activation Block (NCAB) is taken as input and passes through three Self-Attention Enhancement Blocks serially, one after another. The process is described as follows:

\begin{equation}
    F_{\text{SAEB}_1} = \text{SAEB}_1(F_{\text{NCAB}})
\end{equation}

\begin{equation}
    F_{\text{SAEB}_2} = \text{SAEB}_2(F_{\text{SAEB}_1})
\end{equation}

\begin{equation}
    F_{\text{SAEB}_3} = \text{SAEB}_3(F_{\text{SAEB}_2})
\end{equation}

Where, \( F_{\text{SAEB}_3} \) is the output of the Initial TriSAE Module, with 256 output channels.

\subsubsection{QuadSAE Module}
The QuadSAE Module, serving as the secondary module in the architecture, comprises four Self-Attention Enhancement Blocks (SAEB). The output \( F_{\text{SAEB}_3} \) from the Initial TriSAE Module is used as the input for the QuadSAE Module. This input sequentially passes through the four Self-Attention Enhancement Blocks. This Module has 256 input channels and 512 output channels. The process is mathematically represented as follows:

\begin{equation}
    F_{\text{SAEB}_4} = \text{SAEB}_4(F_{\text{SAEB}_3})
\end{equation}

\begin{equation}
    F_{\text{SAEB}_5} = \text{SAEB}_5(F_{\text{SAEB}_4})
\end{equation}

\begin{equation}
    F_{\text{SAEB}_6} = \text{SAEB}_6(F_{\text{SAEB}_5})
\end{equation}

\begin{equation}
    F_{\text{SAEB}_7} = \text{SAEB}_7(F_{\text{SAEB}_6})
\end{equation}

Where, \( F_{\text{SAEB}_7} \) is the output of the QuadSAE Module, with 512 output channels.

\begin{figure*}[ht]
    \centering
    \includegraphics[width=\textwidth]{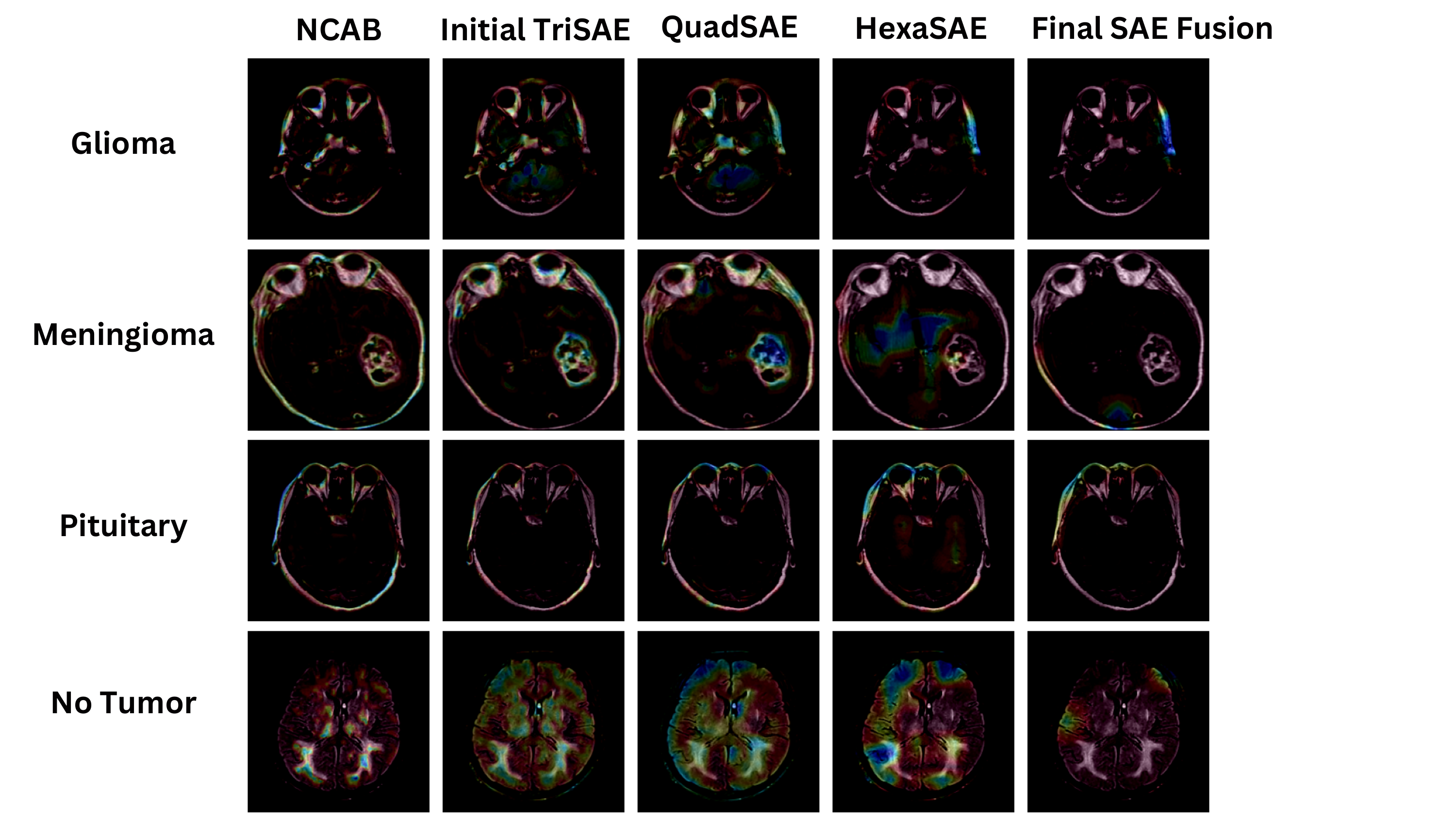}
    \caption{Comparison of all metrics on different models on dataset 1}
    \label{fig: gradcam}
\end{figure*}

\subsubsection{HexaSAE Module}
The Operations of this HexaSAE Module is similar to the Initial TriSAE Module and QuadSAE Module. But the difference is it has six Self Attention Enhancement Modules and the input has 512 channels and 1024 output channels. The output \( F_{\text{SAEB}_7} \) of the QuadSAE module is taken as input and computed by the HexaSAE function to present the ultimate Output of this module. The process of this module is represented in equation \ref{eq:hexasae}.

\begin{equation}
F_{\text{out}} = K_{\text{HexaSAE}}(F_{\text{SAEB}_7})
    \label{eq:hexasae}
\end{equation}

Where, \( F_{\text{out}} \) is the output of the HexaSAE Module, with 1024 output channels and \(K_{\text{HexaSAE}}\) is the six SAE Block Operations performed in HexaSAE Module.

\subsubsection{Final SAE Fusion Module}
The Final SAE Fusion Module is the Final Module of our Deep Learning Architecture that takes 1024 channels as input and produces an output of 2048 channels. It consisted of three Self Attention Enhancement blocks like the TriSAE Module but has a difference in the input-output channels. The output of the HexaSAE Module is taken as input and passes through three SAE blocks to compute the Output. The process is represented as follows:

\begin{equation}
    \text{FSAEFM}_{out} = \text{SAEB}_{16}(\text{SAEB}_{15}(\text{SAEB}_{14}(F_{\text{out}})))
\end{equation}

Where, \( \text{FSAEFM}_{out} \) is the output of the Final SAE Fusion Module, with 2048 output channels and \( F_{\text{out}} \) is the output of the HexaSAE Module.\\

After passing the Final SAE Fusion Module the output goes through an Average Pooling Layer which calculates the average value of each patch of the filter-covered feature map. This operation is beneficial for smoothing and minimizing the effects of outliers \cite{pool1, pool2}. A flattened layer is used to transform the multi-dimensional output of the proposed CNN network into a one-dimensional vector. This transformation is necessary because fully connected (dense) layers, which typically follow convolutional layers, require a one-dimensional input. In dense layer 2048 neorons are assigned for non-linear classification with an activation function ReLU. For a positive input, the ReLU function outputs the value directly; if not, it outputs zero \cite{relu1, relu2}. In the Final Dense Layer, 4 neurons are placed to classify 3 types of tumor and Non-Tumor classes. A softmax activation is used in the final layer which converts logits into probabilities\cite{softmax}.

\begin{figure*}[ht]
    \includegraphics[width=\textwidth]{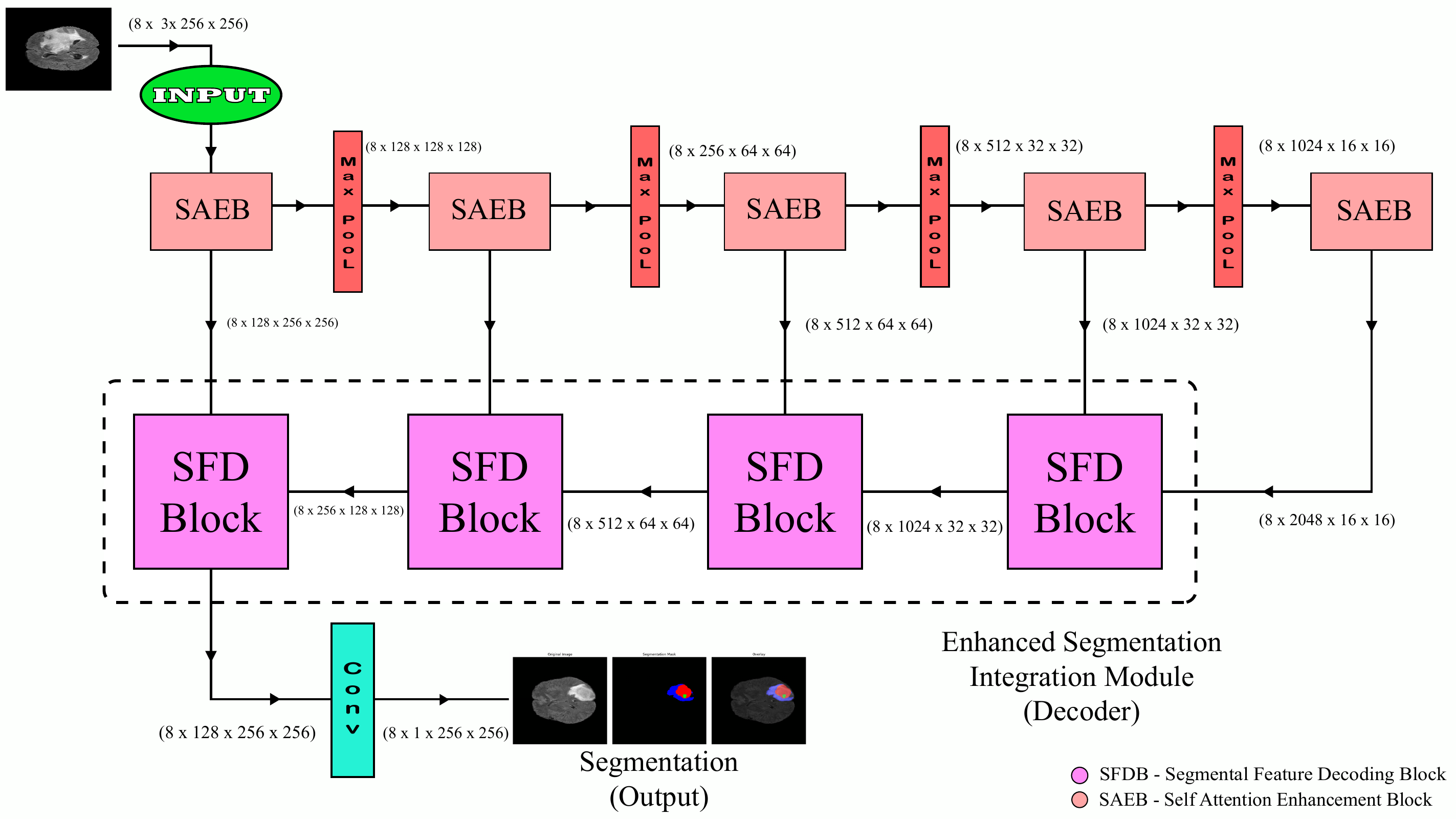}
    \caption{Self-Attentive Segmentation Network}
    \label{fig: sasnetwork}
\end{figure*}

\begin{figure*}[ht]
    \includegraphics[width=\textwidth]{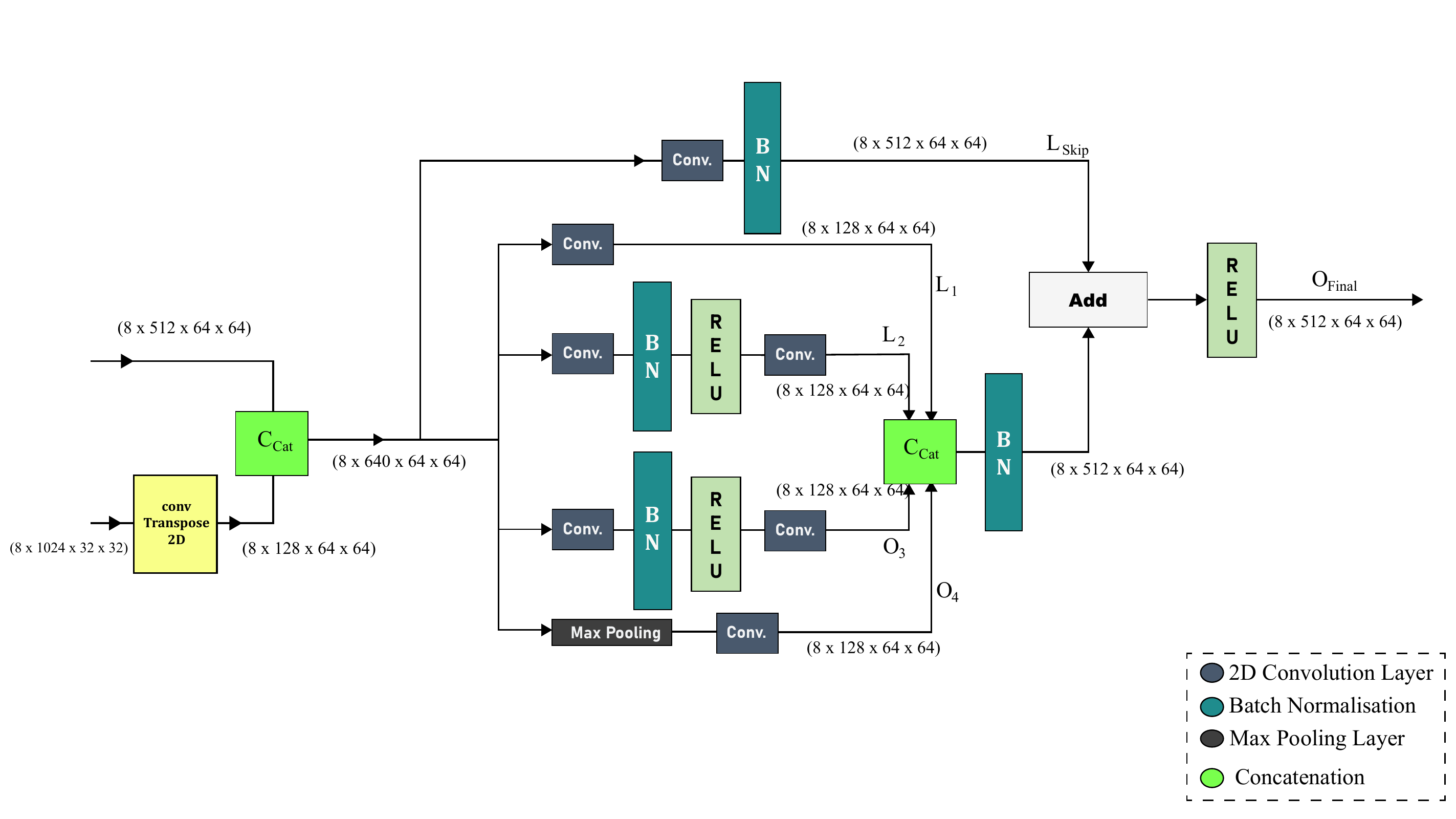}
    \caption{SFD Block}
    \label{fig: sfdblock}
\end{figure*}

\subsubsection{Model Complexity Analysis}

In Saeedi et al. \cite{rw1} a 2D CNN has been proposed whose mathematical equation is represented in the equation \ref{eq:analysis ref}. They have used sequential layers of two convolutions followed by a maxpooling and this has been repeated four times sequentially with different filter sizes (64,32,16 and 8 respectively). 

\begin{equation}
\begin{split}
X_{\text{out}} &= \text{Dropout}(\text{MaxPool}(\text{Conv}(\text{Conv}( \\
&\ldots \text{Dropout}(\text{MaxPool}(\text{Conv}(\text{Conv}(X))))))))) \\
\end{split}
\label{eq:analysis ref}
\end{equation}

As compared to our model they have not used any residual or inception operation which can help in multiple feature extraction and faster convergence. The inception operation used in our model is one of the big reasons for achieving perfect classification accuracy and beating other's works. The main foundation of our proposed architecture is our Self Attention Enhancement Block. We have used four parallel connections and a skip connection. 

The First 1x1 convolution acts as a linear transformation, reducing dimensionality and capturing spatial relationships without changing the feature map size. It retains essential information while making the model computationally efficient. This connection helps in identifying fine-grained, low-level features. 

In the second Connection, The initial 1x1 convolution helps in reducing dimensionality, followed by batch normalization, which normalizes activations, and ReLU adds non-linearity. The second 1x1 convolution further refines the extracted features. This connection is capable of learning more complex features due to the added non-linearity and normalization, ensuring stable and faster convergence.

The third connection works quite a bit the same as the second connection, but its parallel nature allows learning a slightly different set of features due to different initial weights and gradient flows. It provides redundancy and robustness, capturing diverse patterns that might be missed by a single path.

In the fourth connection, the max pooling reduces the spatial dimensions, capturing dominant features and making the subsequent convolution focus on these. The 1x1 convolution then processes this reduced information efficiently. This connection highlights the most prominent features, ensuring that the essential information is retained and further refined.

the fifth connection acts as a skip connection ensuring that the network can learn residuals, thus combating the vanishing gradient problem. This helps in stable and faster training. It Helps in capturing both the initial low-level features and the refined features from the other paths, improving the overall feature richness.

In our model, we have used 4 different modules for extracting different features. The first Module extract the primary edges of the brain along with the edges of the tumor and other critical parts of it. It just extracts the features of the outline, while the second module goes deep into the outline and extracts features on the deep pixel levels. In this module, clear predictions can be made whether the tumor is present or not. The third module is given to understand the pixeled images to classify different types of tumors present in the tumor. This layer analyses the different shapes and structures of different classes of tumors. The fourth module is given to improve performance results based on our experiments. These four are the main building blocks of our proposed model.

The result of our Model is Completely based on repetitive experiments and training step by step and module by module. The only reason to use 16 Self Attention Enhancement Blocks serially one after another, is because this is the optimum value where we are getting the best classification result. The number of SAE Blocks that we add one by one is directly proportional to the best positive results but after a certain number of block( that is 16 in our experiment) the classification ability gets saturated and thus we select that optimum 16 SAE blocks. Those 16 SAE blocks are divided into 4 modules because of their different feature extraction capabilities. This architecture is powerful for capturing the varied and intricate features necessary for accurate brain tumor classification, offering both computational efficiency and improved performance.

\subsection{Segmentation Architecture}
Brain Tumor Segmentation is very crucial for diagnosis planning. Doctors can plan the treatment process by properly localizing the tumor. Thus we have proposed a novel architecture - Self Attentive Segmentation Network, that can accurately segment the tumor region with a pixel accuracy of 99.23\%.

\subsubsection{Self-Attentive Segmentation Network}
 The Self Attentive Segmentation Network consists of five Self Attention Enhancement Blocks connected sequentially with a max pooling layer between them and an Enhanced Segmentation Integration Module as a decoder which has four Segmental Feature Decoding Block in it. Each SAE Block has two outputs. One is connected as an input to the consecutive SAE Block and another is passed as an input to the (k-N)th SFD Block, where k is the total number of SAE Block used in the architecture (here k=5) and N is the current number of the SAE Block. The output from the Enhanced Segmentation Integration Module is subsequently processed through a Convolutional Layer, culminating in the precise segmentation of the tumor. The complete architecture of the Self Attentive Segmentation Network is shown in the figure \ref{fig: sasnetwork}.

\subsubsection{Segmental Feature Decoding Block}
The Segmental Feature Decoder Block (SFD Block) is an integral part of the segmentation model, meticulously designed to enhance the resolution and accuracy of feature maps through a combination of upsampling and feature integration from the encoder. This block begins with a ConvTranspose2d layer, which upsamples the input feature maps, thereby increasing their spatial resolution. Following this, the upsampled features are concatenated with corresponding feature maps from the (k-n)th SAE Blocks(where k is the total number of SAE Block used in the architecture, here k=5 and n is the current number of the SFD Block), ensuring the preservation of critical spatial information. The concatenated feature map then undergoes further processing through a complex Residual Inception module. This module consists of multiple convolutional paths: a 1x1 convolution, a sequence of 1x1 followed by 3x3 convolutions, and a 1x1 followed by 5x5 convolutions, each capturing features at different scales. Additionally, a path involving max pooling followed by a 1x1 convolution is included to capture spatial information from pooled features. The outputs from these paths are concatenated and batch-normalized. Also, there is another skip layer(L-skip) having a 1x1 convolution with batch normalization that gets added to the output of the four parallel layers. And then the added output is activated using ReLU, creating a rich, multi-scale feature representation. A crucial aspect of this block is the residual connection that adds the original input back to the processed output, ensuring efficient gradient flow and mitigating the risk of vanishing gradients. This comprehensive approach results in refined, high-resolution feature maps that significantly enhance the precision of the segmentation, particularly for complex tasks like brain tumor identification. The complete architecture of this block is shown in the figure \ref{fig: sfdblock}. The mathematics is described in detail in the following equations.

\begin{equation}
L_{up} = f_{convT}(X) \tag{1}
\end{equation}

\begin{equation}
L_{concat} = L_{up} \oplus S \tag{2}
\end{equation}

\begin{equation}
L_1 = f_{conv1x1}(L_{concat}) \tag{3}
\end{equation}

\begin{equation}
L_2 = f_{conv3x3}(f_{BN}(f_{ReLU}(f_{conv1x1}(L_{concat})))) \tag{4}
\end{equation}

\begin{equation}
L_3 = f_{conv5x5}(f_{BN}(f_{ReLU}(f_{conv1x1}(L_{concat})))) \tag{5}
\end{equation}

\begin{equation}
L_4 = f_{conv1x1}(f_{pool}(L_{concat})) \tag{6}
\end{equation}

\begin{equation}
L_{skip} = f_{BN}(f_{conv1x1}(L_{concat})) \tag{7}
\end{equation}

\begin{equation}
L_{output} = f_{ReLU}(f_{BN}(L_1 \oplus L_2 \oplus L_3 \oplus L_4) + L_{skip}) \tag{8}
\end{equation}\\

Where,
\( L_{up} \) is the upsampled feature map obtained using the transposed convolution.
\( f_{convT} \) is transposed convolution operation, which upsamples the input feature map \( X \).
\( X \) is the input feature map to the UpBlock.
\( L_{concat} \): The concatenated feature map from upsampling and the skip connection.
\( \oplus \)is concatenation operation.
\( S \)is the skip connection from the corresponding encoder layer.
\( L_1 \) is output of the 1x1 convolution branch.
\( f_{conv1x1} \) is 1x1 convolution operation.
\( L_2 \) is the Output of the 1x1 convolution followed by a 3x3 convolution branch.
\( f_{BN} \) is the Batch normalization function.
\( f_{ReLU} \) is ReLU activation function.
\( f_{conv3x3} \) is the 3x3 convolution operation.
\( L_3 \)is the Output of the 1x1 convolution followed by a 5x5 convolution branch.
\( f_{conv5x5} \) is 5x5 convolution operation.
\( L_4 \) is the Output of the max pooling followed by a 1x1 convolution branch.
\( f_{pool} \) is Max pooling operation.
\( L_{skip} \)is the Output of the shortcut connection.
\( L_{output} \) is the Final output of the Residual Inception Block after combining all branches and adding the residual connection.

\begin{table*}[t]
\centering
\fontsize{8}{15}\selectfont
\begin{tabular}{llllllllllllll}
\hline
\multicolumn{4}{c}{\textbf{Dataset 1}} &           & \multicolumn{4}{c}{\textbf{Dataset 2}} & \multicolumn{1}{c}{} & \multicolumn{4}{c}{\textbf{Dataset 3}} \\ \hline
\textbf{Class} & \textbf{Images} & \textbf{Train} & \textbf{Test} & \textbf{} & \textbf{Class} & \textbf{Images} & \textbf{Train} & \textbf{Test} & \textbf{} & \textbf{Class} & \textbf{Images} & \textbf{Train} & \textbf{Test} \\ \hline
\textit{\textbf{Gliomas}} & 1621 & 1321 & 300 &           & \textit{\textbf{Gliomas}} & 1426 & 1153 & 273 &  & \textit{\textbf{Gliomas}} & 1426 & 1129 & 297 \\ \hline
\textit{\textbf{Meningiomas}} & 1645 & 1339 & 306 &           & \textit{\textbf{Meningiomas}} & 708 & 566 & 142 &  & \textit{\textbf{Meningiomas}} & 708 & 572 & 136 \\ \hline
\textit{\textbf{Pituitary}} & 1757 & 1457 & 300 &           & \textit{\textbf{Pituitary}} & 930 & 732 & 198 &  & \textit{\textbf{Pituitary}} & 930 & 749 & 181 \\ \hline
\textit{\textbf{No Tumor}} & 2000 & 1595 & 405 &           & \textbf{} &  &  &  &  & \textit{\textbf{No Tumor}} & 1485 & 1189 & 296 \\ \hline
\textbf{Total} & \textbf{7023} & \textbf{5712} & \textbf{1311} &           & \textbf{Total} & \textbf{3064} & \textbf{2451} & \textbf{613} &  & \textbf{Total} & \textbf{4549} & \textbf{3639} & \textbf{910} \\ \hline
\end{tabular}
\caption{Dataset Description}
\label{table:dataset}
\end{table*}

\begin{figure}[H]
    \centering
    \includegraphics[width=0.95\columnwidth]{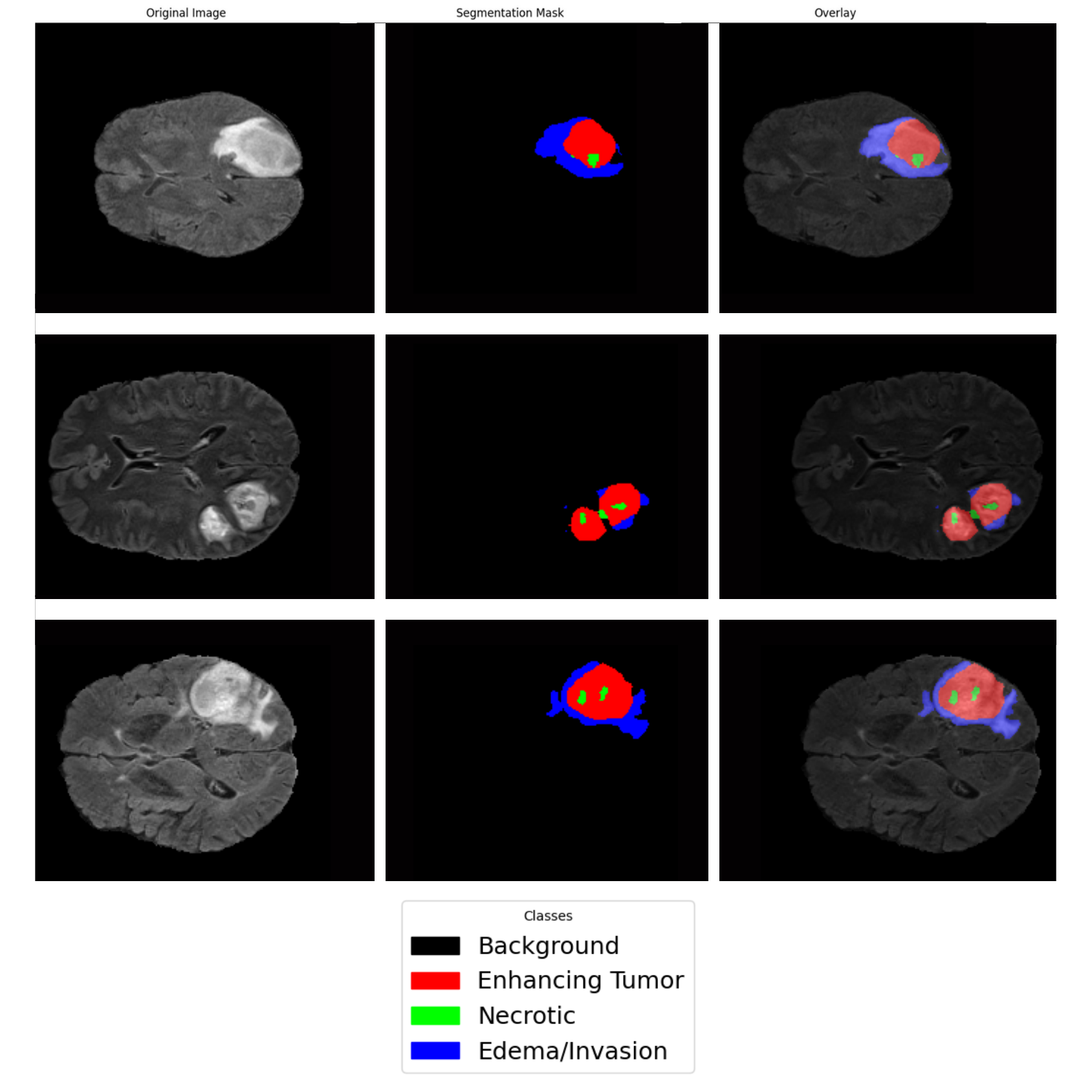}
    \caption{Segmentating the Tumor region accurately}
    \label{fig: segmentation result}
\end{figure}

\subsubsection{Enhanced Segmentation Integration Module}
The Enhanced Segmentation Integration Module acts as a decoder in this novel Architecture. This module consists of four sequential Segmental Feature Decoding Block. The mathematical Operation of this Module is described in the following equation.

\begin{equation} IN(SFD_n) = \begin{cases} SAEB_5 \oplus SAEB_{k-n} & \text{if } n = 1 \\ SFD_{n-1} \oplus SAEB_{k-n} & \text{if } n > 1 \end{cases} 
\label{eq: ESIM}
\end{equation}\\

We define \( SFD_n \) as the \( n \)th SFD block, \( SAEB_k \) as the \( k \)-th SAEB block, and \( IN(SFD_n) \) as the input to the \( n \)-th SFD block. Here, \( \oplus \) represents the concatenation operation. The equation \ref{eq: ESIM} captures the input to each SFD block. For \( n = 1 \), The first SFD block takes inputs from \( SAEB_5 \) and \( SAEB_4 \). For \( n > 1 \), Each subsequent SFD block takes input from the previous SFD block and the corresponding SAEB block (\( k - n \)). Here, k=5, because we have used five SAEB blocks in our architecture.

\section{Experiment and Result}
In this section we have discussed about the dataset and the preprocessing steps, the total experimental setup, and the evaluation metrics along with the results. Also, we have compared our model with previous works and other state-of-the-art existing models.

\subsection{Dataset and Preprocessing}
We have used three publicly available MRI Image datasets for the classification of the Brain tumors in this study. The datasets are as follows:

Brain MRI Dataset (Dataset 1) from Kaggle. This dataset contains 7023 images of human brain MRI images which are classified into four classes - Gliomas(1621), meningiomas(1645), pituitary(1757), and No Tumor(2000). This dataset is a combination of Figshare, SARTAJ, and Br35H datasets. (Reference dataset word 1). The dataset is split into training and test sets in the ratio 80:20. Out of 7023 images in the dataset, 5712 images are taken in the training set and the rest are taken for the test set.

Figshare dataset (Dataset 2) given by Jun Cheng, School of Biomedical Engineering, Southern Medical University, Guangzhou, China. This brain tumor dataset contains 3064 T1-weighted contrast-enhanced images from 233 patients with three kinds of brain tumor: meningioma (708 slices), glioma (1426 slices), and pituitary tumor (930 slices). This data is organized in Matlab data format (.mat file). Each file stores a struct containing the following labels for an image - 1 for meningioma, 2 for glioma, and 3 for pituitary tumor. This is a legitimate dataset used in many researches across the globe. Out of 3064 images in the dataset, 2451 images are chosen to train the network, and the rest 613 images are taken to evaluate the performance of the model. Thus splitting the dataset in the ratio 80:20. (Reference dataset word 2)

The third dataset (Dataset 3) is a custom-made cross-mixed dataset of Dataset 1 and Dataset 2. In this, we have made a custom dataset containing four classes since this is a mixed dataset so a good accurate result in this dataset proves the generalization capability of our proposed architecture. This dataset contains a total of 4549 images out of which 3639 images are taken to train the model and the rest are used to evaluate the performance of the model.

All the details of the datasets and their divisions (i.e., training and testing) are listed in Table \ref{table:dataset} and Visualisation is provided in Figure \ref{fig: total}, \ref{fig: train} and \ref{fig: test}. Dataset 1 and Dataset 3 contain images of different sizes so they are resized to images of similar size. All the images are resized to (224x224) pixels size and then basic preprocessing steps like contrast enhancement and normalization are performed. On the other hand, the Figshare dataset (Dataset 2) contains images in MAT files in raw form. So preprocessing steps are performed to convert the images from .mat files to .jpg files to use for training and test set to train and evaluate our network respectively. After conversion to image files, the Dataset is split into training and test sets in a ratio of 80:10. Other preprocessing steps that are performed are z-score normalization and rescaling pixel values to the specified range(0,1).




We have used the BratS2020 dataset for Segmentation. The BRATS2020 dataset is a comprehensive resource for brain tumor segmentation, specifically focusing on gliomas. It includes multi-institutional, pre-operative MRI scans from 19 different institutions, encompassing T1, T1 post-contrast, T2, and T2-FLAIR modalities2. The dataset provides ground truth labels for the enhancing tumor, peritumoral edema, and necrotic/non-enhancing tumor core, manually segmented by expert neuroradiologists. This rich, multimodal dataset is designed to advance the development and evaluation of segmentation algorithms, with the ultimate goal of improving brain tumor diagnosis and treatment planning.



\subsection{Experimental Setup}
\begin{figure}[H]
    \centering
    \begin{subfigure}[b]{\columnwidth}
        \centering
        \includegraphics[width=\columnwidth]{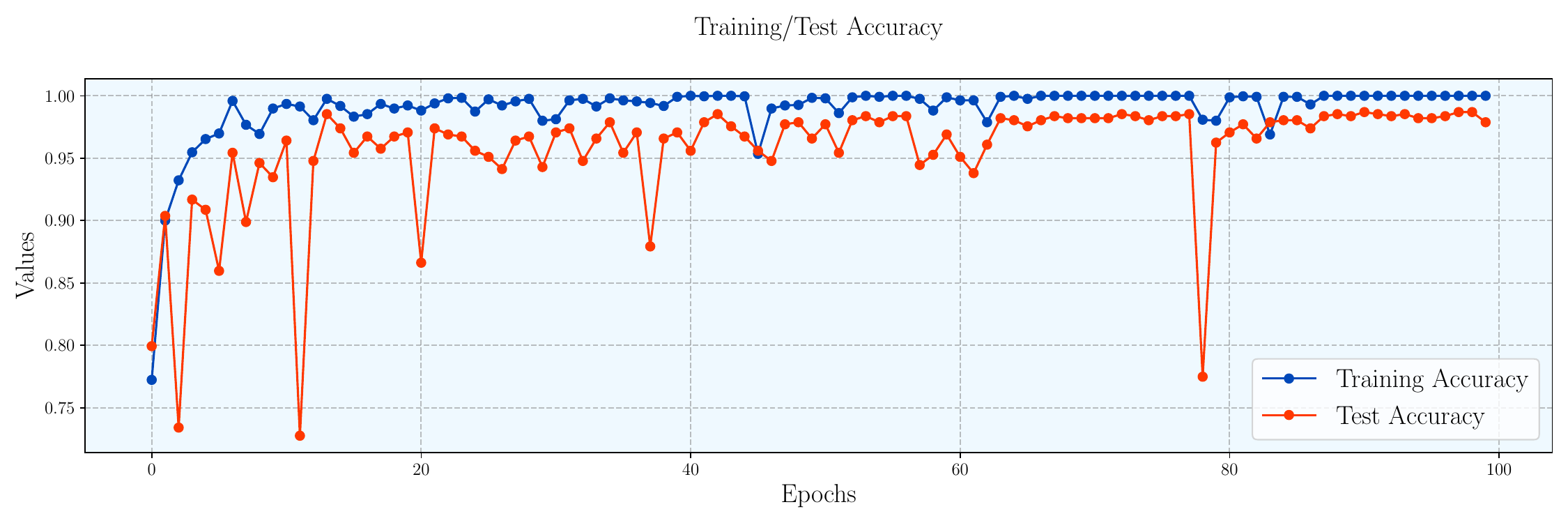}
        \caption{}
    \end{subfigure}
    \hfill
    \begin{subfigure}[b]{\columnwidth}
        \centering
        \includegraphics[width=\columnwidth]{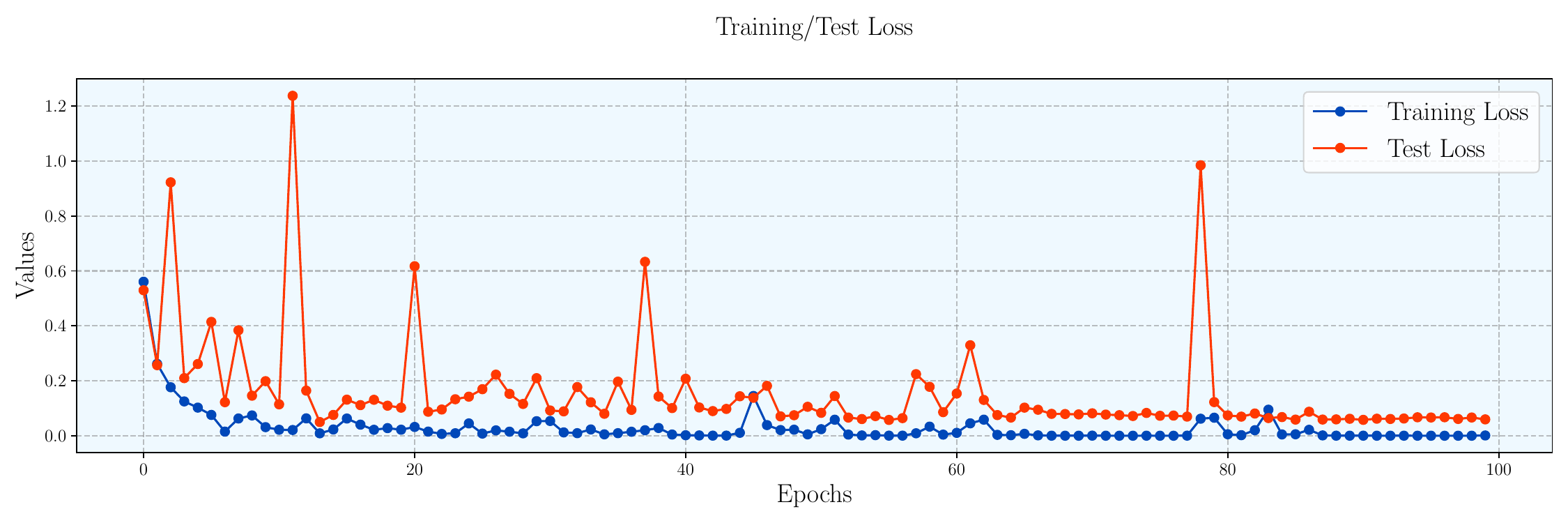}
        \caption{}
    \end{subfigure}
    \captionsetup{justification=centering}
    \caption{Accuracy and Loss Metrics for Dataset 2 (Figshare). (a) Accuracy and (b) Loss}
    \label{fig: figshare accloss}
\end{figure}

\begin{figure}[H]
    \centering
    \begin{subfigure}[b]{\columnwidth}
        \centering
        \includegraphics[width=\columnwidth]{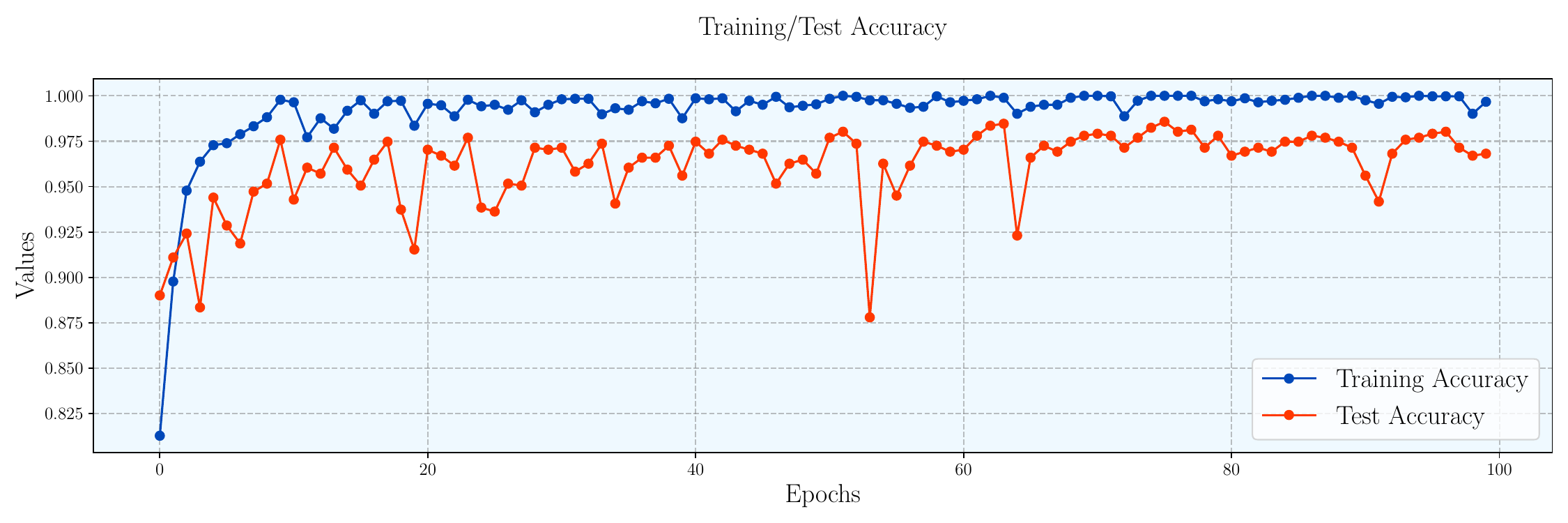}
        \caption{}
    \end{subfigure}
    \hfill
    \begin{subfigure}[b]{\columnwidth}
        \centering
        \includegraphics[width=\columnwidth]{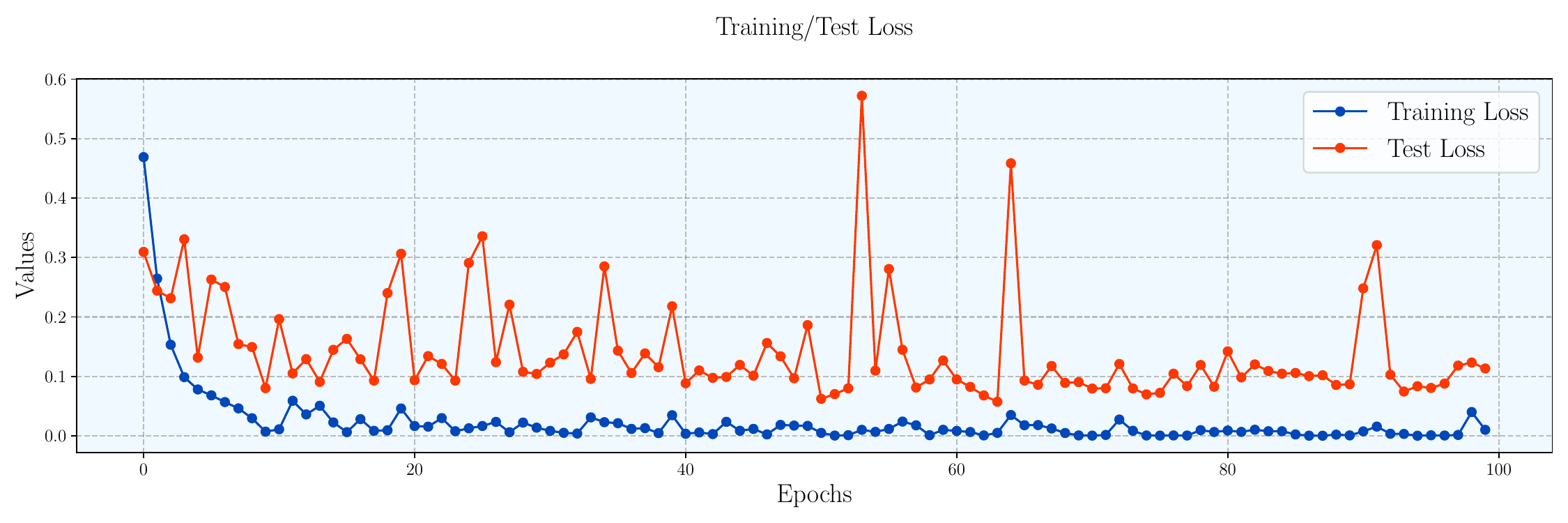}
        \caption{}
    \end{subfigure}
    \captionsetup{justification=centering}
    \caption{Accuracy and Loss Metrics for Dataset 3 (Custom Dataset). (a) Accuracy and (b) Loss}
    \label{fig: custom accloss}
\end{figure}

Many images are of different shapes so we have to resize those images in patches of 224x224 and then they are preprocessed. All the datasets are split into training and test sets in the ratio of 80:20 before passing the training set to train the neural network for classification. We have used a learning rate of 0.0001, and the "Adam" Optimizer with a cross-entropy Loss function for classification and Binary Cross Entropy with Logits Loss for Segmentation purposes, to train our model. The Loss functions are represented in the equation \ref{loss func} and \ref{eq: bceloss}. The accuracy and loss graph plot is shown in Figure \ref{fig: figshare accloss} and \ref{fig: custom accloss}, and the confusion matrix of the test dataset of different datasets are shown in Figure \ref{fig: cmd1}, \ref{fig: cmd2} and \ref{fig: cmd3}. Our architecture mainly includes an NCA block and 16 SAE blocks divided into four Modules - Initial TriSAE, QuadSAE, HexaSAE, and Final TriSAE Fusion Module as shown in the figure \ref{fig: saetcn}. The proposed model is implemented in pytorch. The hardware specifications and Software environment used by us to develop this proposed model and to train it are represented in Table \ref{table:specifications}. The process flow diagram of the experimental process is shown in figure \ref{fig: processflow}.

\begin{equation}
    L = -\sum_{i=1}^{N} \sum_{c=1}^{C} y_{i,c} \log(p_{i,c})
    \label{loss func}
\end{equation}

Where,
\( N \) is the number of samples,
\( C \) is the number of classes,
\( y_{i,c} \) is a binary indicator (0 or 1) if class label \( c \) is the correct classification for sample \( i \),
\( p_{i,c} \) is the predicted probability that sample \( i \) is of class \( c \).

\begin{equation} \label{eq: bceloss} \text{BCE} = -\frac{1}{N} \sum_{i=1}^{N} \left[ y_i \log(p_i) + (1 - y_i) \log(1 - p_i) \right] \end{equation}

Where, \( N \) is the number of samples. \( y_i \) is the true label for the \( i \)-th sample (either 0 or 1). \( p_i \) is the predicted probability for the \( i \)-th sample. \( \log \) is the natural logarithm function. \( 1 - y_i \) is the complement of the true label for the \( i \)-th sample. \( 1 - p_i \) is the complement of the predicted probability for the \( i \)-th sample. \( \sum \) denotes summation over all \( N \) samples. \( -\frac{1}{N} \) normalizes the loss by dividing the total loss by the number of samples.

\begin{figure}[H]
    \centering
    \includegraphics[width=0.95\columnwidth]{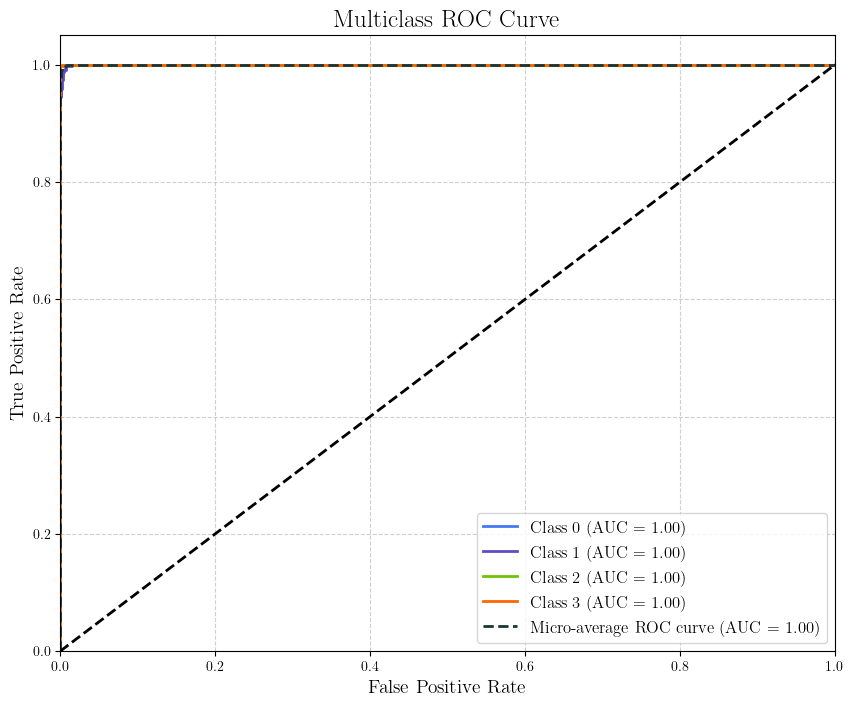}
    \captionsetup{justification=centering}
    \caption{ROC-AUC Curve of our proposed model on the dataset 1}
    \label{fig:roc-auc}
\end{figure}

\begin{table}[H]
\centering
\fontsize{10}{17}\selectfont
\begin{tabular}{ll}
\hline
\multicolumn{2}{|c|}{\textbf{Hardware Specifications}} \\ \hline
CPU & \textit{Ryzen 7 7700x} \\ \hline
GPU & \textit{Rtx 4060ti 8GB} \\ \hline
RAM & \textit{16 GB} \\ \hline
\multicolumn{2}{|c|}{\textbf{Software Environment}} \\ \hline
Operating System & \textit{Windows 11} \\ \hline
Programming Languages & \textit{Python} \\ \hline
Libraries & \textit{Pytorch, Scikit Learn} \\ \hline
\end{tabular}
\caption{Hardware Specifications and Software Environment used in our Study}
\label{table:specifications}
\end{table}

\subsection{Evaluation Metrics for Classification}
It is very much essential to evaluate the performances of a proposed architecture for image classification purposes to investigate the proper functioning of the deep learning model in real-life scenarios. In this study, we have used different standard parametric methods for evaluation of our model like precision, F1 Score, R2 Score, Recall, confusion Matrix, and ROC-AUC curve. For multiclass classification there are two kinds of Metrics evaluation, one is Macro Averaged Metrics and the other is Micro Averaged Metrics. In Macro Averaged Metrics, it calculates the metric independently for each class and then takes the average (treating all classes equally). On the other hand, Micro-averaging aggregates the contributions of all classes to compute the average metric (treating all instances equally). In this Study we have evaluated the Micro Averaged Metrics. These are some standard evaluation metrics which are discussed one by one along with their mathematical representations:

\begin{itemize}

    \item \textbf{Precision}: Precision measures the accuracy of the positive predictions.
    \begin{equation}
    \text{Micro Precision} = \frac{\sum_{i=1}^{N} TP_i}{\sum_{i=1}^{N} (TP_i + FP_i)}
    \end{equation}
    
    \item \textbf{Recall}: Recall measures the ability of the model to find all the relevant cases.
    \begin{equation}
    \text{Micro Recall} = \frac{\sum_{i=1}^{N} TP_i}{\sum_{i=1}^{N} (TP_i + FN_i)}
    \end{equation}

    \item \textbf{F1 Score}: The F1 Score is the harmonic mean of precision and recall.
    \begin{equation}
    \text{Micro F1 Score} = 2 \cdot \frac{\text{Micro Precision} \cdot \text{Micro Recall}}{\text{Micro Precision} + \text{Micro Recall}}
    \end{equation}

    \item \textbf{R2 Score (Coefficient of Determination)}: The R2 Score indicates how well the model's predictions approximate the real data points.
    \begin{equation}
    R^2 = 1 - \frac{\sum_{i=1}^{N} (y_i - \hat{y}_i)^2}{\sum_{i=1}^{N} (y_i - \bar{y})^2}
    \end{equation}


    \item \textbf{ROC-AUC Curve (Receiver Operating Characteristic - Area Under Curve)}: The ROC curve is a graphical representation of the true positive rate (TPR) against the false positive rate (FPR) at various threshold settings. The AUC represents the degree or measure of separability. The ROC-AUC curve of our study is shown in figure \ref{fig:roc-auc} Here we have used One vs One (OvO) Approach (refer to equation \ref{eq:ovo}) and Micro Averaged Approach (refer to equation \ref{eq:micro}) to calculate the ROC-AUC. The expressions are showed as follows:
    \begin{equation}
    \text{AUC}_{\text{OvO}} = \frac{2}{N(N-1)} \sum_{i=1}^{N-1} \sum_{j=i+1}^{N} \text{AUC}_{ij}
    \label{eq:ovo}
    \end{equation}
    \begin{equation}
    \text{AUC}_{\text{micro}} = \frac{\sum_{i=1}^{N} \sum_{j=1}^{N} \text{AUC}_{ij}}{N^2}
    \label{eq:micro}
    \end{equation}

\end{itemize}

In this section, the notation used for evaluation metrics includes the following: \( N \) is the total number of classes. \( TP_i \) represents the True Positives for class \( i \), which are the instances correctly predicted as class \( i \). \( FP_i \) denotes the False Positives for class \( i \), which are the instances incorrectly predicted as class \( i \). \( FN_i \) stands for the False Negatives for class \( i \), which are the instances that belong to class \( i \) but were incorrectly predicted as another class. \( TN_i \) indicates the True Negatives for class \( i \), which are the instances correctly predicted as not belonging to class \( i \). \( y_i \) is the actual value for instance \( i \), while \( \hat{y}_i \) is the predicted value for instance \( i \). \( \bar{y} \) represents the mean of the actual values. Additionally, \( N \) is the total number of classes, and \( \text{AUC}_{ij} \) is the AUC score for the binary classification problem between class \( i \) and class \( j \), all of which are used in ROC-AUC calculations.

\subsection{Evaluation Metrics for Segmentation}
When evaluating segmentation models, several key metrics are essential to gauge their performance accurately. Intersection Over Union (IoU) measures the overlap between the predicted segmentation and the ground truth, providing a ratio of their intersection over their union. The Matthews Correlation Coefficient (MCC) offers a balanced measure of binary classification, considering true and false positives and negatives, making it particularly useful even when class sizes differ. The Dice Similarity Coefficient (DSC), assesses the similarity between the predicted and actual segments by comparing their overlap relative to their sizes. Specificity evaluates how well the model identifies true negatives, which is crucial for understanding the model's ability to recognize non-tumorous regions. Finally, the Boundary F1 Score (BF1) focuses on the accuracy of the predicted boundaries of the segmentation, ensuring that the model can precisely delineate the edges of the segmented objects. Together, these metrics provide a comprehensive understanding of the segmentation model's accuracy, robustness, and reliability. All of these are described below on eby one with their mathematical equations.

\begin{itemize} \item \textbf{Intersection Over Union (IoU)} \begin{equation} IoU_c = \frac{TP_c}{TP_c + FP_c + FN_c} \end{equation} IoU measures the overlap between the predicted segmentation and the ground truth. It is calculated as the area of overlap divided by the area of union between the predicted and ground truth masks. 

\item \textbf{Matthews Correlation Coefficient (MCC)} \\

MCC is a balanced measure that takes into account true and false positives and negatives and is generally regarded as a balanced metric for binary classification, even if the classes are of very different sizes. 

\item \textbf{Dice Similarity Coefficient (DSC)} \begin{equation} DSC_c = \frac{2 \cdot TP_c}{2 \cdot TP_c + FP_c + FN_c} \end{equation} The DSC, also known as the F1 Score, measures the similarity between two sets. It is calculated as twice the area of overlap divided by the total number of pixels in both the predicted and ground truth masks. \item \textbf{Specificity} \begin{equation} Specificity_c = \frac{TN_c}{TN_c + FP_c} \end{equation} Specificity measures the proportion of true negatives that are correctly identified. It is calculated as the number of true negatives divided by the sum of true negatives and false positives. \item \textbf{Boundary F1 Score (BF1)} \begin{equation} BF1_c = \frac{2 \cdot |P_c \cap G_c|}{|P_c| + |G_c|} \end{equation} The Boundary F1 Score evaluates the accuracy of boundary prediction. It is similar to the DSC but focuses on the boundaries of the segmented objects. \end{itemize}

\begin{table}[H]
\centering
\fontsize{8}{15}\selectfont
\begin{tabular}{ccccccc}
\hline
        &  & \textbf{Accuracy} & \textbf{Precision} & \textbf{Recall} & \textbf{F1 Score} & \textbf{ROC AUC} \\ \hline
\textbf{Class 0} &  & 99.79    & 99.89     & 99.89  & 99.89    & 99.98   \\ \hline
\textbf{Class 1} &  & 59.01    & 69.10     & 74.19  & 71.76    & 99.41   \\ \hline
\textbf{Class 2} &  & 82.27    & 84.71     & 83.24  & 83.97    & 99.94   \\ \hline
\textbf{Class 3} &  & 84.89    & 91.32     & 91.32  & 89.71    & 99.98   \\ \hline
\end{tabular}
\caption{Class wise Evaluation metrics of Segmentation}
\label{tab:evaluation segmentation}
\end{table}

\begin{table}[H]
\centering
\fontsize{8}{15}\selectfont
\begin{tabular}{ccccccc}
\hline
        &  & \textbf{IoU}   & \textbf{Specificity} & \textbf{MCC}   & \textbf{Boundary F1 Score} & \textbf{DSC}   \\ \hline
\textbf{Class 0} &  & 99.58 & 93.55       & 93.41 &                   & 99.79 \\ \hline
\textbf{Class 1} &  & 45.85 & 99.86       & 71.47 & 89.53             & 62.87 \\ \hline
\textbf{Class 2} &  & 69.72 & 99.90       & 83.86 & 92.45             & 82.16 \\ \hline
\textbf{Class 3} &  & 71.87 & 99.95       & 89.67 & 97.90             & 83.64 \\ \hline
\end{tabular}
\caption{Class wise Evaluation}
\label{tab:special evaluation segmentation}
\end{table}

\subsection{Performance of our Proposed Classification Model on individual datasets}
\begin{figure}[H]
    \centering
    \includegraphics[width=0.95\columnwidth]{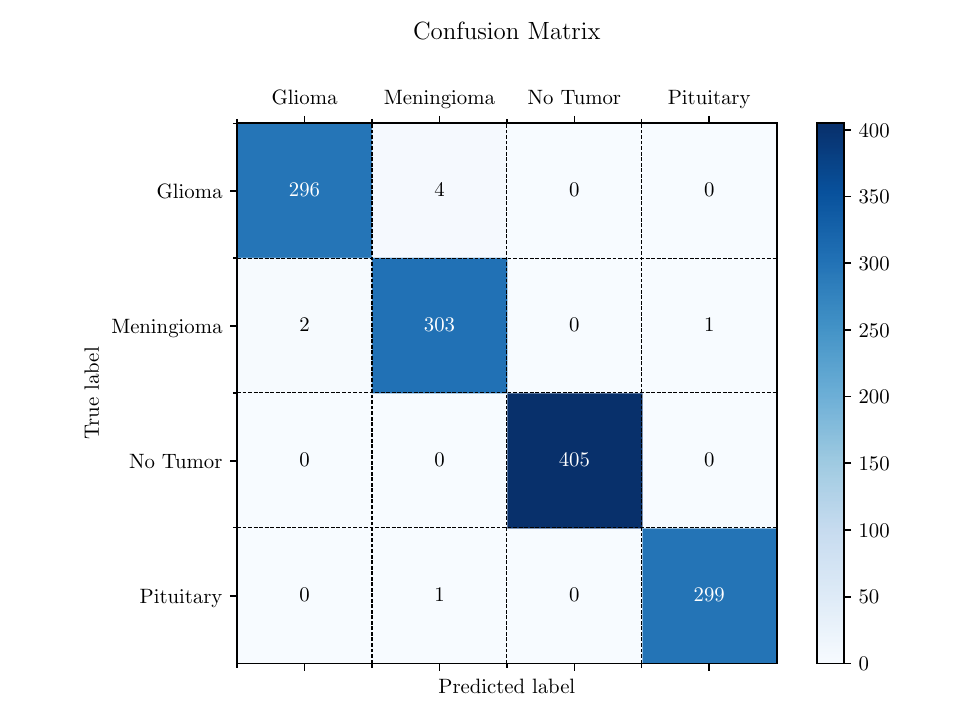}
    \caption{Confusion Matrix of Dataset 1}
    \label{fig: cmd1}
\end{figure}

\begin{figure}[H]
    \centering
    \includegraphics[width=0.95\columnwidth]{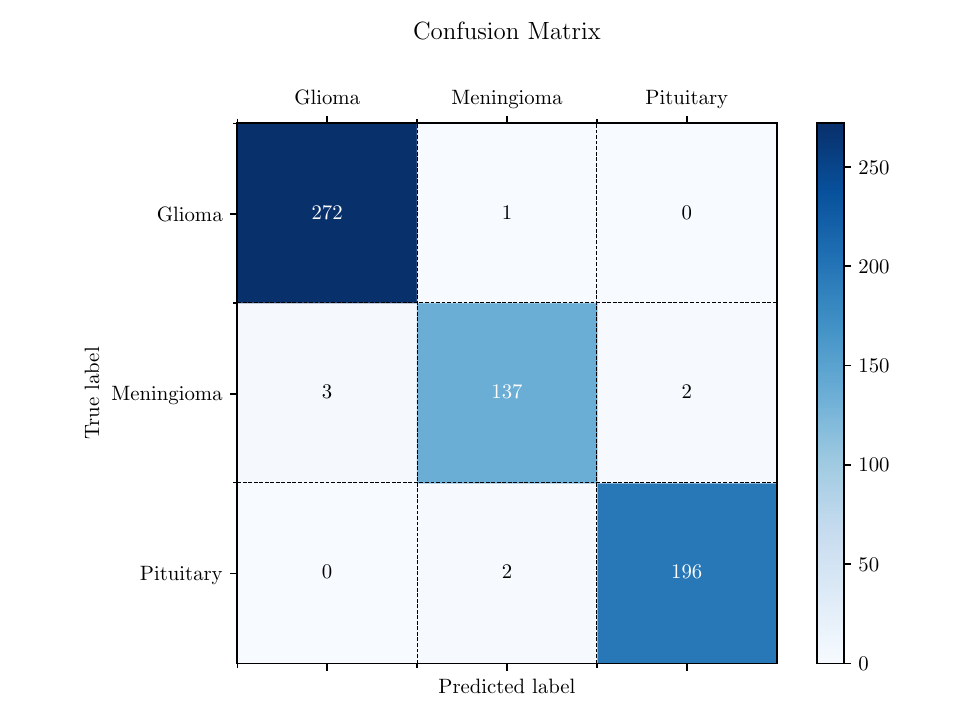}
    \caption{Confusion Matrix of Dataset 2}
    \label{fig: cmd2}
\end{figure}

\begin{figure}[H]
    \centering
    \includegraphics[width=0.95\columnwidth]{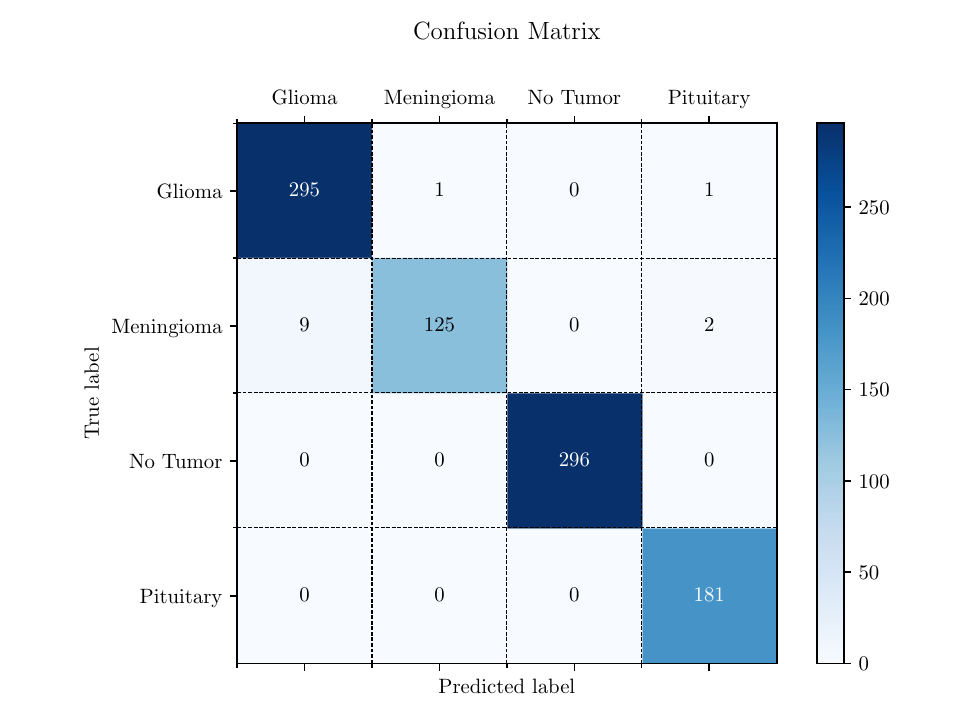}
    \caption{Confusion Matrix of Dataset 3}
    \label{fig: cmd3}
\end{figure}

We have trained our proposed network on three datasets and their results are compared in the table \ref{table:dataset comparison} and the confusion matrix of the test data of those three datasets are shown in the figure \ref{fig: cmd1}, \ref{fig: cmd2}, \ref{fig: cmd3}.
\vspace{2cm}

\begin{table*}[ht]
\centering
\fontsize{8}{15}\selectfont
\begin{tabular}{|c|c|c|c|c|}
\hline
\textbf{Contribution} & \textbf{Year of Publish} & \textbf{Types of Classifier} & \textbf{Dataset} & \textbf{Test Accuracy} \\ \hline
\textit{Muhammad Aamir et al. \cite{rw5}} & 2024 & Hyperparametric CNN & Brain MRI (Kaggle) & 97\% \\ \hline
 &  & 2D CNN &  & 93.44\% \\ \cline{3-3} \cline{5-5} 
\multirow{-2}{*}{\textit{Saeedi et al. \cite{rw1}}} & \multirow{-2}{*}{2023} & Auto-Encoder & \multirow{-2}{*}{\begin{tabular}[c]{@{}c@{}}Brain Tumor Classification \\ (MRI): Four Classes\end{tabular}} & 90.92\% \\ \hline
 &  &  & \begin{tabular}[c]{@{}c@{}}Figshare and Multiclass\\ Kaggle Dataset (Augmented)\end{tabular} & 98.12\% \\ \cline{4-5} 
\multirow{-2}{*}{\textit{Takowa Rahman et al. \cite{rw4}}} & \multirow{-2}{*}{2023} & \multirow{-2}{*}{PDCNN} & \begin{tabular}[c]{@{}c@{}}Figshare and Multiclass\\ Kaggle Dataset (Original)\end{tabular} & 95.60\% \\ \hline
\textit{Badza et al. \cite{rw7}} & 2020 & 22 layers CNN & Figshare Dataset & 96.56\% \\ \hline
\textit{Abiwinanda et al. \cite{rw2}} & 2018 & CNN & Figshare Dataset & 84.19\% \\ \hline
 &  & DCNet &  & 93.04\% \\ \cline{3-3} \cline{5-5} 
\multirow{-2}{*}{\textit{Sai Samarth R. Phaye et al. \cite{rw6}}} & \multirow{-2}{*}{2018} & DCNet++ & \multirow{-2}{*}{Figshare Dataset} & 95.03\% \\ \hline
\textit{Pashaei et al. \cite{rw8}} & 2018 & CNN & Figshare Dataset & 93.68\% \\ \hline
\textit{Paul et al. \cite{rw9}} & 2017 & CNN & Figshare Dataset & 91.43\% \\ \hline
 &  &  & \textbf{Brain MRI (Kaggle)} & \textbf{99.38\%} \\ \cline{4-5} 
 &  &  & \textbf{Figshare Dataset} & \textbf{98.69\%} \\ \cline{4-5} 
\multirow{-3}{*}{\textit{\textbf{Our Proposed Architecture}}} & \multirow{-3}{*}{\textbf{2024}} & \multirow{-3}{*}{\textbf{\begin{tabular}[c]{@{}c@{}}SAETCN - Self-Attention\\ Enhancement Tumor \\ Classification Network\end{tabular}}} & \textbf{\begin{tabular}[c]{@{}c@{}}Cross-Mixed Custom\\ Dataset\end{tabular}} & \textbf{98.57\%} \\ \hline
\end{tabular}
\caption{Comparison of Our Work (Classification Architecture - SAETCN) with State of the Art Works}
\label{table:my-previous work comparison}
\end{table*}

\begin{table}[H]
\centering
\fontsize{6}{15}\selectfont
\begin{tabular}{lcccccc}
\hline
\textbf{Name of the Dataset} & \textbf{Accuracy} & \textbf{Precision} & \textbf{F1 Score} & \textbf{Recall} & \textbf{R2 Score} & \textbf{ROC-AUC} \\ \hline
\textit{Dataset 1 (Brain MRI )} &
  \textbf{99.38} &
  \textbf{99.39} &
  \textbf{99.39} &
  \textbf{99.38} &
  \textbf{99.08} &
  \textbf{99.99} \\ \hline
\textit{Dataset 2 (Figshare)} &
  98.69 &
  98.69 &
  98.68 &
  98.69 &
  98.25 &
  99.90 \\ \hline
\textit{\begin{tabular}[c]{@{}l@{}}Dataset 3 (Custom \\ Cross-Mixed)\end{tabular}} &
  98.57 &
  98.59 &
  98.55 &
  98.57 &
  97.69 &
  99.90 \\ \hline
\end{tabular}
\caption{Comparison of our model on Different Datasets}
\label{tab:dataset comparison}
\end{table}

\subsection{Comparison of our Classification Model with Previous Works}
We have compared our model with various state-of-the-art models like EfficientNetB4, ResNet18, InceptionNetV3, Swin Transformer, and ViT (Vission Transformer) but ultimately we have beat all those models by achieving an accuracy of 99.38\%, 98.69\% and 98.57\% in dataset 1 dataset 2 and dataset 3 respectively (refer to table \ref{table:dataset comparison}).
All the evaluation metrics of all the models compared with our model on different datasets are also shown in detail in table\ref{table:comparison dataset1}, table\ref{table:comparison dataset2}, table \ref{table:comparison dataset3}. In table \ref{table:my-previous work comparison} we have compared our proposed model with other previous works that was performed in the last recent years.
\vspace{2cm}

\begin{table}[H]
\centering
\fontsize{6.5}{15}\selectfont
\begin{tabular}{lcccccc}
\hline
\multicolumn{1}{c}{\multirow{2}{*}{\textbf{Model}}} & \multicolumn{6}{c}{\textbf{Dataset 1}}                                                                      \\ \cline{2-7} 
\multicolumn{1}{c}{}                                & \textbf{} & \textbf{Precision} & \textbf{Recall} & \textbf{F1 Score} & \textbf{R2 Score} & \textbf{ROC-AUC} \\ \hline
EfficientNetB4   &           & 95.31 & 95.31 & 95.21 & 92.82 & 99.46 \\ \hline
ResNet18         &           & 98.22 & 98.22 & 98.22 & 96.09 & 99.91 \\ \hline
InceptionNetV3   &           & 98.95 & 98.90 & 98.94 & 98.25 & 99.97 \\ \hline
Swin Transformer &           & 94.41 & 94.43 & 94.41 & 84.86 & 99.27 \\ \hline
ViT              & \textbf{} & 87.57 & 87.56 & 87.51 & 67.95 & 96.85 \\ \hline
\textbf{Proposed}                                   &           & \textbf{99.39}     & \textbf{99.38}  & \textbf{99.39}    & \textbf{99.08}    & \textbf{99.99}   \\ \hline
\end{tabular}
\caption{Comparison of our Proposed model with Other state of the art existing Models in terms of Evaluation Metrics on Brain MRI Dataset (Dataset 1)}
\label{tab:comparison dataset1}
\end{table}

\begin{table}[H]
\centering
\fontsize{6.5}{15}\selectfont
\begin{tabular}{lcccccc}
\hline
\multicolumn{1}{c}{\multirow{2}{*}{\textbf{Model}}} & \multicolumn{6}{c}{\textbf{Dataset 2}}                                                                      \\ \cline{2-7} 
\multicolumn{1}{c}{}                                & \textbf{} & \textbf{Precision} & \textbf{Recall} & \textbf{F1 Score} & \textbf{R2 Score} & \textbf{ROC-AUC} \\ \hline
EfficientNetB4   &           & 91.73 & 91.68 & 91.70 & 84.47 & 97.66 \\ \hline
ResNet18         &           & 96.22 & 96.24 & 96.22 & 94.29 & 99.49 \\ \hline
InceptionNetV3   &           & 97.87 & 97.87 & 97.85 & 97.19 & 99.85 \\ \hline
Swin Transformer &           & 87.85 & 87.92 & 87.87 & 72.75 & 96.03 \\ \hline
ViT              & \textbf{} & 79.00 & 79.77 & 79.26 & 59.84 & 90.25 \\ \hline
\textbf{Proposed}                                   &           & \textbf{98.69}     & \textbf{98.69}  & \textbf{98.68}    & \textbf{98.25}    & \textbf{99.90}  
\end{tabular}
\caption{Comparison of our Proposed model with Other state of the art existing Models in terms of Evaluation Metrics on Figshare Dataset (Dataset 2)}
\label{tab:comparison dataset2}
\end{table}

\begin{table}[H]
\centering
\fontsize{6.5}{15}\selectfont
\begin{tabular}{lcccccc}
\hline
\multicolumn{1}{c}{\multirow{2}{*}{\textbf{Model}}} & \multicolumn{6}{c}{\textbf{Dataset 3}}                                                                      \\ \cline{2-7} 
\multicolumn{1}{c}{}                                & \textbf{} & \textbf{Precision} & \textbf{Recall} & \textbf{F1 Score} & \textbf{R2 Score} & \textbf{ROC-AUC} \\ \hline
EfficientNetB4   &           & 96.16 & 96.15 & 96.14 & 90.53 & 99.51 \\ \hline
ResNet18         &           & 97.12 & 97.14 & 97.13 & 96.19 & 99.80 \\ \hline
InceptionNetV3   &           & 98.05 & 98.02 & 98.03 & 95.42 & 99.89 \\ \hline
Swin Transformer &           & 92.82 & 92.96 & 92.86 & 89.69 & 98.81 \\ \hline
ViT              & \textbf{} & 85.38 & 85.60 & 85.47 & 69.55 & 96.52 \\ \hline
\textbf{Proposed}                                   &           & \textbf{98.59}     & \textbf{98.57}  & \textbf{98.55}    & \textbf{97.69}    & \textbf{99.90}  
\end{tabular}
\caption{Comparison of our Proposed model with Other State of the art existing Models in Terms of Evaluation Metrics on Cross-Mixed Dataset (Dataset 3)}
\label{tab:comparison dataset3}
\end{table}


\begin{table*}[ht]
\centering
\fontsize{8}{15}\selectfont
\begin{tabular}{cccccccccccc}
\hline
\textbf{NCA} & \textbf{\begin{tabular}[c]{@{}c@{}}Initial\\ TriSAE\end{tabular}} & \textbf{QuadSAE} & \textbf{HexaSAE} & \textbf{\begin{tabular}[c]{@{}c@{}}Final SAE\\ Fusion\end{tabular}} & \textbf{} & \textbf{Accuracy} & \textbf{F1} & \textbf{R2} & \textbf{Recall} & \textbf{Precision} & \textbf{AUC} \\
\hline
\ding{51} & & & & & & 59.42 & 59.94 & -0.31 & 59.42 & 63.28 & 85.34 \\
\hline
\ding{51} & \ding{51} & & & & & 83.68 & 83.82 & 39.57 & 83.67 & 87.27 & 98.80 \\
\hline
\ding{51} & \ding{51} & \ding{51} & & & & 98.09 & 98.09 & 98.02 & 98.09 & 98.13 & 99.94 \\
\hline
\ding{51} & \ding{51} & \ding{51} & \ding{51} & & & 96.26 & 96.28 & 95.28 & 96.26 & 96.40 & 99.88 \\
\hline
\ding{51} & \ding{51} & \ding{51} & \ding{51} & \ding{51} & & 99.38 & 99.39 & 99.08 & 99.38 & 99.39 & 99.99 \\
\hline
\end{tabular}
\caption{Module Wise Ablation Study of the Proposed Model components}
\label{tab:abalation study}
\end{table*}



\begin{table}[H]
\centering
\fontsize{6.5}{15}\selectfont
\begin{tabular}{lccccccccc}
\hline
\multicolumn{1}{c}{\multirow{2}{*}{\textbf{Model}}} &
   &
  \multicolumn{2}{c}{\textbf{Dataset 1}} &
   &
  \multicolumn{2}{c}{\textbf{Dataset 2}} &
   &
  \multicolumn{2}{c}{\textbf{Dataset 3}} \\ \cline{2-10} 
\multicolumn{1}{c}{} &  & \textbf{Train} & \textbf{Test}  &  & \textbf{Train} & \textbf{Test}  &  & \textbf{Train} & \textbf{Test}  \\ \hline
EfficientNetB4       &  & 99.57          & 95.31          &  & 98.98          & 91.68          &  & 99.98          & 96.15          \\ \hline
ResNet18             &  & 99.90          & 98.22          &  & 99.85          & 96.24          &  & 99.56          & 97.14          \\ \hline
InceptionNetV3       &  & 100            & 98.90          &  & 99.20          & 97.87          &  & 100            & 98.02          \\ \hline
Swin Transformer     &  & 97.25          & 94.43          &  & 98.29          & 87.92          &  & 99.84          & 92.96          \\ \hline
ViT                  &  & 96.56          & 87.56          &  & 96.24          & 79.77          &  & 97.89          & 85.60          \\ \hline
\textbf{Proposed}    &  & \textbf{1.00}  & \textbf{99.38} &  & \textbf{99.89} & \textbf{98.69} &  & \textbf{99.46} & \textbf{98.57} \\ \hline
\end{tabular}
\caption{Comparison of Our Proposed Model with State of the art models between Dataset 1 (Brain MRI), Dataset 2 (Figshare) and Dataset 3 (Custom Cross Mixed Dataset) with respect to Train and Test accuracy}
\label{tab:dataset comparison}
\end{table}

\subsection{Comparison of our segmentation model with previous works}
Table \ref{tab:segmentation comparison} presents a comparative analysis of the proposed SAS-Net architecture against previous works by Agarwal et al. and Wentau Wu et al. The SAS-Net outperforms the other models across most metrics, achieving a Dice Similarity Coefficient (DSC) of 99.79\%, specificity of 93.55\%, and sensitivity of 99.89\%. Notably, while Wentau Wu et al.'s model exhibits high specificity at 99.82\%, it falls behind in DSC and sensitivity compared to SAS-Net. Meanwhile, Agarwal et al.'s model shows a balanced performance but with lower sensitivity at 79.89\%. Overall, the SAS-Net demonstrates superior performance, particularly in terms of DSC and sensitivity, indicating its effectiveness in accurately segmenting brain tumors.

\begin{table}[H]
\centering
\fontsize{6.5}{15}\selectfont
\begin{tabular}{cccccc}
\hline
                            &  & \textbf{DSC} & \textbf{Specificity} & \textbf{Sensitivity} & \textbf{Accuracy} \\ \hline
\textbf{Agarwal et al.} \cite{aggarwal2023early}    &  & 94.5         & 92.6                 & 79.89                & 91.3              \\ \hline
\textbf{Wentau Wu et al.}\cite{wu2020intelligent}   &  & 89.58        & 99.82                & 91.10                & -                 \\ \hline
\textbf{SAS-Net (Proposed)} &  & 99.79        & 93.55                & 99.89                & 99.23             \\ \hline
\end{tabular}
\caption{Comparison of our proposed segmentation architecture with previous works}
\label{tab:segmentation comparison}
\end{table}

\section{Abalation Study}
\begin{figure}[h]
    \centering
    \includegraphics[width=\columnwidth]{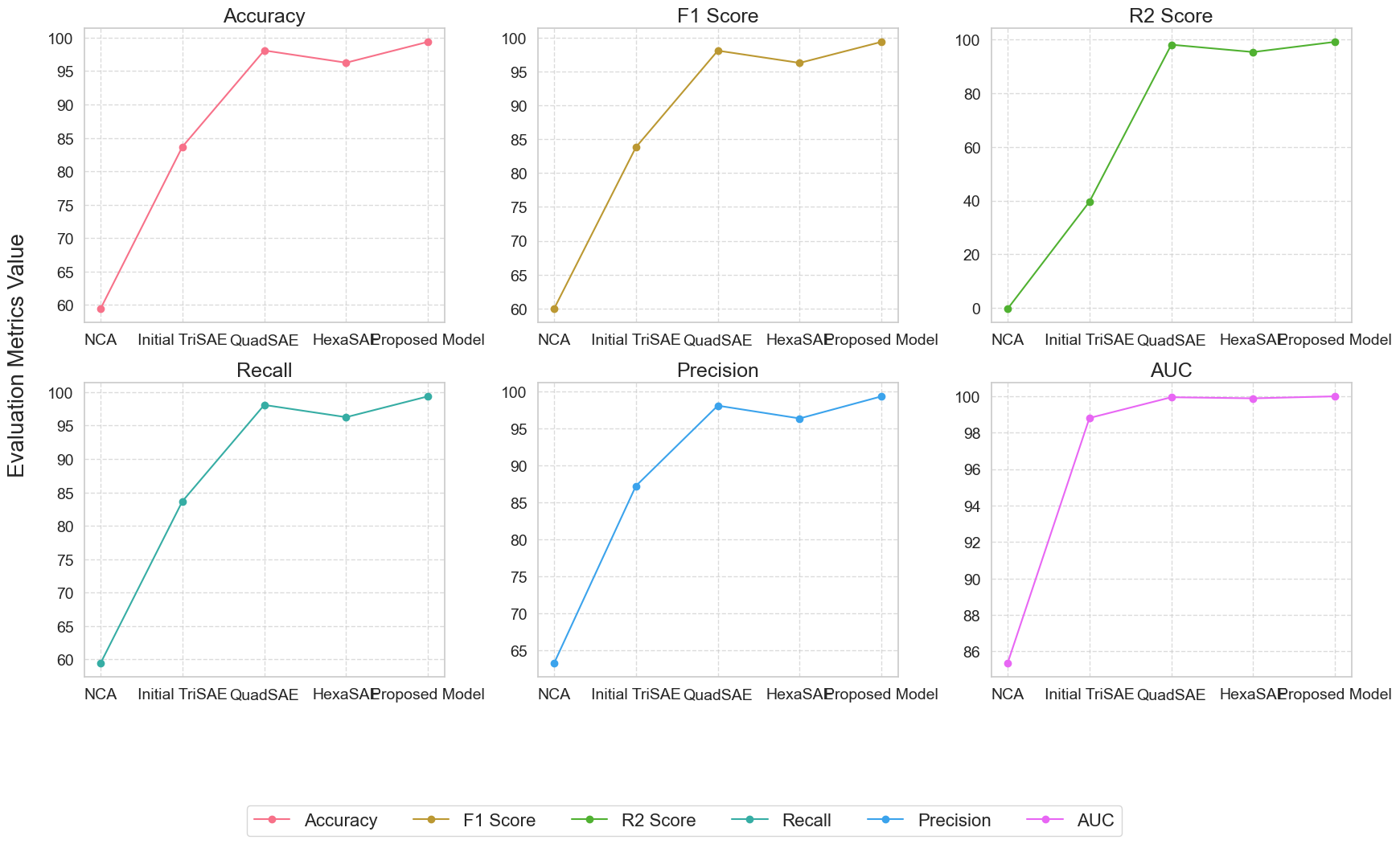}
    \caption{Representation of the evaluation metrics on different model components}
    \label{fig: ablation study plot}
\end{figure}

Module wise ablation study is performed and evaluated. The result is shown in the table \ref{tab:abalation study}. In figure \ref{fig: ablation study plot}, the improvement of the result is represented that describes the view of choosing 16 SAEB blocks as this gives the better classification performance amongst every combinations. A detailed Gradient-weighted Class Activation Mapping of all model components is shown in figure \ref{fig: gradcam}.


\section{Conclusions}
\label{sec:conclusions}
In this paper, we proposed SAETCN - Self-Attention Enhancement Tumor Classification Network for detecting and classification of three kinds of Brain tumors. This study explored methods for improving the accuracy of detection and classification of different kinds of Brain Tumor - Gliomas, Meningiomas, and Pituitary. Our model can also predict whether the brain has any tumor or not. Since Brain Tumor is a huge problem and detecting and classifying those require a lot of time for radiologists and is hectic and thus hinders the proper treatment plan and diagnosis. Therefore this type of advanced CAD system is very much required in the Medical field. Manual Classification also leads to human error but with an accuracy of 99.31\% the Artificial Intelligence Model produces less error than manual classification. Also, it can be integrated with Mobile or web applications which increases the ease of use and thus people in remote areas who are not getting proper healthcare also use our web-based model to detect whether they are suffering from a brain tumor or not. We have compared our model with other state-of-the-art models and with other researchers' previous works, we found that our model has beaten most of the model by achieving an accuracy of 99.31\%, 99.20\%, and in dataset 1, dataset 2, and dataset 3 respectively. Although our proposed model has achieved great accuracy, still there is a probability that the model may get overfitted on the three datasets we have used in this study, thus it should be trained in larger datasets before using it for real-world classification in Hospitals. More research and development can be done in this work in the future with more accurate and generalized results and thus in this way, artificial intelligence will help mankind to be better in the upcoming days.

Although the accuracy of our proposed model is quite high still there is a lot of future scope for this study, they include making the model more generalized for working more accurately in real-time classification. This can be achieved by training the network in Big data with a machine with high computational resources to train it effectively. Also, in the future, we have planned to integrate this model in a Mobile application, where doctors and patients can be the users and detect the brain tumor and classify it with their proper location thus helping in the treatment. If they want they can also use their own data to train the model and fine-tune it based on their specific MRI machine to get better results of classification.

\section{Data Availablity}
\label{sec:data availablity}

For maximum transparency and to foster open science, all computational codes and saved model architectures related to this study have been made publicly available in the full GitHub repository at \url{https://github.com/arghadip2002/SAETCN-and-SASNET-Architectures}. 

The datasets utilized, which include two publicly available datasets and a custom-made dataset, are linked within the repository (\url{https://github.com/arghadip2002/SAETCN-and-SASNET-Architectures/blob/main/dataLinks}), with the Custom dataset accessible upon reasonable request to the author. 

Furthermore, the practical utility of the novel SAETCN and SASNET architectures is demonstrated by their integration into the backend of the \href{https://huggingface.co/spaces/arghadip2002/NeuroGuard-Web-Application}{NeuroGuard Web Application}, which performs classification and detection of various brain tumors from MRI images, thereby showcasing the direct application of Artificial Intelligence in the medical domain for building future-ready applications.

\bibliographystyle{plain}
\bibliography{mybib}

\begin{thebibliography}{10}

\bibitem{rw5}
Muhammad Aamir, Abdallah Namoun, Sehrish Munir, Nasser Aljohani, Meshari~Huwaytim Alanazi, Yaser Alsahafi, and Faris Alotibi.
\newblock Brain tumor detection and classification using an optimized convolutional neural network.
\newblock {\em Diagnostics}, 14(16):1714, 2024.

\bibitem{rw2}
Nyoman Abiwinanda, Muhammad Hanif, S~Tafwida Hesaputra, Astri Handayani, and Tati~Rajab Mengko.
\newblock Brain tumor classification using convolutional neural network.
\newblock In {\em World Congress on Medical Physics and Biomedical Engineering 2018: June 3-8, 2018, Prague, Czech Republic (Vol. 1)}, pages 183--189. Springer, 2019.

\bibitem{aggarwal2023early}
Mukul Aggarwal, Amod~Kumar Tiwari, M~Partha Sarathi, and Anchit Bijalwan.
\newblock An early detection and segmentation of brain tumor using deep neural network.
\newblock {\em BMC Medical Informatics and Decision Making}, 23(1):78, 2023.

\bibitem{rw7}
Milica~M Bad{\v{z}}a and Marko~{\v{C}} Barjaktarovi{\'c}.
\newblock Classification of brain tumors from mri images using a convolutional neural network.
\newblock {\em Applied Sciences}, 10(6):1999, 2020.

\bibitem{bouhafra2024deep}
Sara Bouhafra and Hassan El~Bahi.
\newblock Deep learning approaches for brain tumor detection and classification using mri images (2020 to 2024): A systematic review.
\newblock {\em Journal of Imaging Informatics in Medicine}, pages 1--31, 2024.

\bibitem{i5}
Cleveland Clinic.
\newblock Brain cancer (brain tumor).
\newblock \url{https://my.clevelandclinic.org/health/diseases/6149-brain-cancer-brain-tumor}, 2022.

\bibitem{softmax}
Michael Franke and Judith Degen.
\newblock The softmax function: Properties, motivation, and interpretation.
\newblock 2023.

\bibitem{i4}
Anna Giorgi.
\newblock What is a brain tumor?
\newblock \url{https://www.verywellhealth.com/brain-tumor-7253734}, April 06, 2023.
\newblock You can access the data online on the provided link.

\bibitem{relu2}
Richard~HR Hahnloser, Rahul Sarpeshkar, Misha~A Mahowald, Rodney~J Douglas, and H~Sebastian Seung.
\newblock Correction: Digital selection and analogue amplification coexist in a cortex-inspired silicon circuit.
\newblock {\em Nature}, 408(6815):1012--1012, 2000.

\bibitem{i1}
National~Cancer Institute.
\newblock Cancer stat facts: Brain and other nervous system cancer.
\newblock \url{https://seer.cancer.gov/statfacts/html/brain.html}, 2014-2020.

\bibitem{relu1}
Alex Krizhevsky, Ilya Sutskever, and Geoffrey~E Hinton.
\newblock Imagenet classification with deep convolutional neural networks.
\newblock {\em Advances in neural information processing systems}, 25, 2012.

\bibitem{pool2}
Yann LeCun, L{\'e}on Bottou, Yoshua Bengio, and Patrick Haffner.
\newblock Gradient-based learning applied to document recognition.
\newblock {\em Proceedings of the IEEE}, 86(11):2278--2324, 1998.

\bibitem{i3}
Dongdong Lin, Ming Wang, Yan Chen, Jie Gong, Liang Chen, Xiaoyong Shi, Fujun Lan, Zhongliang Chen, Tao Xiong, Hu~Sun, et~al.
\newblock Trends in intracranial glioma incidence and mortality in the united states, 1975-2018.
\newblock {\em Frontiers in oncology}, 11:748061, 2021.

\bibitem{liu2023deep}
Zhihua Liu, Lei Tong, Long Chen, Zheheng Jiang, Feixiang Zhou, Qianni Zhang, Xiangrong Zhang, Yaochu Jin, and Huiyu Zhou.
\newblock Deep learning based brain tumor segmentation: a survey.
\newblock {\em Complex \& intelligent systems}, 9(1):1001--1026, 2023.

\bibitem{rw8}
Ali Pashaei, Hedieh Sajedi, and Niloofar Jazayeri.
\newblock Brain tumor classification via convolutional neural network and extreme learning machines.
\newblock In {\em 2018 8th International conference on computer and knowledge engineering (ICCKE)}, pages 314--319. IEEE, 2018.

\bibitem{rw9}
Justin~S Paul, Andrew~J Plassard, Bennett~A Landman, and Daniel Fabbri.
\newblock Deep learning for brain tumor classification.
\newblock In {\em Medical Imaging 2017: Biomedical Applications in Molecular, Structural, and Functional Imaging}, volume 10137, pages 253--268. SPIE, 2017.

\bibitem{rw6}
Sai Samarth~R Phaye, Apoorva Sikka, Abhinav Dhall, and Deepti Bathula.
\newblock Dense and diverse capsule networks: Making the capsules learn better.
\newblock {\em arXiv preprint arXiv:1805.04001}, 2018.

\bibitem{rw4}
Takowa Rahman and Md~Saiful Islam.
\newblock Mri brain tumor detection and classification using parallel deep convolutional neural networks.
\newblock {\em Measurement: Sensors}, 26:100694, 2023.

\bibitem{i2}
Neuro-Oncology Branch Scientific~Communications Raleigh~McElvery.
\newblock Statistical report highlights key trends in adolescents and young adults with brain tumors.
\newblock \url{https://shorturl.at/MhBCT}, 2024.
\newblock National Cancer Institute.

\bibitem{rao2021comprehensive}
Champakamala~Sundar Rao and K~Karunakara.
\newblock A comprehensive review on brain tumor segmentation and classification of mri images.
\newblock {\em Multimedia Tools and Applications}, 80(12):17611--17643, 2021.

\bibitem{rw1}
Soheila Saeedi, Sorayya Rezayi, Hamidreza Keshavarz, and Sharareh R.~Niakan~Kalhori.
\newblock Mri-based brain tumor detection using convolutional deep learning methods and chosen machine learning techniques.
\newblock {\em BMC Medical Informatics and Decision Making}, 23(1):16, 2023.

\bibitem{i6}
{The Johns Hopkins University}, {The Johns Hopkins Hospital}, and {Johns Hopkins Health System}.
\newblock Brain tumors and brain cancer.
\newblock \url{https://www.hopkinsmedicine.org/health/conditions-and-diseases/brain-tumor}, 2024.

\bibitem{verma2024comprehensive}
Amit Verma, Shiv~Naresh Shivhare, Shailendra~P Singh, Naween Kumar, and Anand Nayyar.
\newblock Comprehensive review on mri-based brain tumor segmentation: A comparative study from 2017 onwards.
\newblock {\em Archives of Computational Methods in Engineering}, pages 1--47, 2024.

\bibitem{wu2020intelligent}
Wentao Wu, Daning Li, Jiaoyang Du, Xiangyu Gao, Wen Gu, Fanfan Zhao, Xiaojie Feng, and Hong Yan.
\newblock An intelligent diagnosis method of brain mri tumor segmentation using deep convolutional neural network and svm algorithm.
\newblock {\em Computational and mathematical methods in medicine}, 2020(1):6789306, 2020.

\bibitem{pool1}
Matthew~D Zeiler and Rob Fergus.
\newblock Visualizing and understanding convolutional networks.
\newblock In {\em Computer Vision--ECCV 2014: 13th European Conference, Zurich, Switzerland, September 6-12, 2014, Proceedings, Part I 13}, pages 818--833. Springer, 2014.

\end{thebibliography}

\end{document}